\documentclass{article}

\usepackage[table,x11names,dvipsnames,svgnames]{xcolor}
\usepackage[final]{corl_2026}
\hypersetup{
pdftitle={MPC-Injection: Biasing Off-Policy Locomotion RL Toward Controller-Induced Behavior Basins},
pdfauthor={Roy Xing, Seyoung Ree, Brian Plancher},
pdfkeywords={reinforcement learning, model predictive control, legged locomotion}
}
\usepackage{subcaption}
\usepackage{amsmath,amsfonts,amssymb}
\providecommand{\mathbbold}[1]{\mathbb{#1}}

\usepackage{fix-cm}

\usepackage{nag}

\makeatletter
\@ifpackageloaded{xcolor}{}{%
\usepackage[table,x11names,dvipsnames,svgnames]{xcolor}%
}
\makeatother

\usepackage{colortbl}

\usepackage{graphicx}
\usepackage{wrapfig}

\definecolorset{RGB}{lyft}{}{Red,194,39,36;Sunset,202,53,33;Orange,205,68,20;Amber,200,117,42;Yellow,242,169,52;Citron,186,188,44;Lime,112,159,33;Green,56,139,31;Mint,45,118,56;Teal,52,133,135;Cyan,60,132,202;Blue,55,94,248;Indigo,64,13,247;Purple,115,42,248;Pink,176,25,145;Rose,176,32,75}

\usepackage{microtype}

\usepackage[american]{babel}

\usepackage{array}
\usepackage{multirow}
\usepackage{booktabs}
\usepackage{makecell} 

\ifcsname labelindent\endcsname

\fi
\usepackage[inline]{enumitem}

\newenvironment{lenumerate}[2][]
{\begin{enumerate}[label=(#2\arabic*),leftmargin=0.2in,itemindent=0.15in,#1]}
{\end{enumerate}}

\setlist*[enumerate,1]{label={\itshape\arabic*)}}

\makeatletter
\newcommand{\paragraphswithstop}{%
\let\copyparagraph\paragraph%
\renewcommand\paragraph[1]{\copyparagraph{##1.}}%
}
\makeatother

\usepackage[framemethod=tikz]{mdframed}

\makeatletter
\def\namedlabel#1#2{\begingroup
  #2%
  \def\@currentlabel{#2}%
  \phantomsection\label{#1}\endgroup
}
\makeatother

\makeatletter
\def\namedlabelphantom#1#2{\begingroup
  \def\@currentlabel{#2}%
  \phantomsection\label{#1}\endgroup
}
\makeatother

\newcommand{\parunskip}{\bgroup\unskip\parfillskip=0pt \par\egroup}

\input{preamble/math}

\usepackage{units}

  \newcommand{\newcolorlabel}[2]{%
  \expandafter\newcommand\csname #1\endcsname[1]{%
    \tikz[baseline]{\node[text=white,fill=#2,anchor=base,text height=1.3ex,text depth=0.1ex,font=\sffamily\bfseries]{##1}}}%
}

\newcommand{\newcommenter}[2]{%
  \expandafter\newcommand\csname #1\endcsname[1]{%
    \fcolorbox{#2}{#2}{\color{white}\textsf{\textbf{#1}}}
    {\color{#2}##1}}%
  \expandafter\newcommand\csname at#1\endcsname{%
    \fcolorbox{#2}{#2}{\color{white}\textsf{\textbf{@#1}}}
    {\color{#2}}}%
  \expandafter\newcommand\csname #1cite\endcsname[1]{%
    \csname #1\endcsname {[##1]}
  }%
  \expandafter\newcommand\csname #1ref\endcsname[1]{%
    \csname #1\endcsname {$\blacktriangleright$##1}
  }%
  \expandafter\newcommand\csname #1hl\endcsname[2]{%
    \colorbox{#2}{\color{white}\textsf{\textbf{#1}}}\sethlcolor{Azure2}\hl{##2}~%
    \expandafter\ifx\csname commentarrow\endcsname\relax$\leftarrow$\else \commentarrow[#2]\fi~%
    {\color{#2}##1}}%
  \expandafter\newcommand\csname #1st\endcsname[2]{%
    \colorbox{#2}{\color{white}\textsf{\textbf{#1}}}\sout{##2}~%
    \expandafter\ifx\csname commentarrow\endcsname\relax$\leftarrow$\else \commentarrow[#2]\fi~%
    {\color{#2}##1}}%
}
\newcommenter{TODO}{DodgerBlue1}
\newcommenter{rtron}{Green3}

\usepackage{comment}

\usepackage{soul}

\usepackage[normalem]{ulem}

\usepackage{csquotes}

\usepackage{suffix}

\usepackage{environ}

\makeatletter
\newsavebox{\boxifnotempty}
\newcommand{\displayifnotempty}[3]{\sbox\boxifnotempty{#2}\setbox0=\hbox{\usebox{\boxifnotempty}\unskip}%
  \ifdim\wd0=0pt
  \else
  #1\usebox{\boxifnotempty}#3%
  \fi%
}

\newcommand{\ifempty}[2]{\setbox0=\hbox{#1\unskip}%
  \ifdim\wd0=0pt%
  #2%
  \fi%
}

\newcommand{\ifnotempty}[2]{\setbox0=\hbox{#1\unskip}%
  \ifdim\wd0>0pt%
  #2%
  \fi%
}
\makeatother

\newcommand{\switchifempty}[3]{\sbox\boxifnotempty{#1}\setbox0=\hbox{\usebox{\boxifnotempty}\unskip}%
  \ifdim\wd0=0pt{}%
  #2%
  \else{}%
  #3%
  \usebox{\boxifnotempty}%
  \fi%
}

\makeatletter
\@ifundefined{chapter}{\usepackage{algorithm}}{\usepackage[chapter]{algorithm}}
\makeatother
\usepackage{algorithmicx}
\usepackage{algpseudocode}
\makeatother%

\usepackage{scrlfile}

\makeatletter
\newcommand*\newstoreddef[1]{
  \BeforeClosingMainAux{%
    \immediate\write\@auxout{%
      \string\restoredef{#1}{\csname #1\endcsname}%
    }%
  }%
}
\newcommand*{\restoredef}[2]{
  \expandafter\gdef\csname stored@#1\endcsname{#2}%
}
\newcommand*{\storeddef}[1]{
  \@ifundefined{stored@#1}{0}{\csname stored@#1\endcsname}%
}
\makeatother

\usepackage{pageslts}
\NewEnviron{tee}{\BODY\typeout{Marker Tee [start] ^^J \BODY ^^JMaker Tee [end]}}

\usepackage{cleveref}

\usepackage{tikz}
\usetikzlibrary{calc}
\usetikzlibrary{matrix}
\usetikzlibrary{chains,scopes}
\usetikzlibrary{shapes.geometric}
\usetikzlibrary{arrows.meta}
\usetikzlibrary{decorations.markings}
\usetikzlibrary{decorations.pathreplacing}
\usetikzlibrary{backgrounds}

\tikzset{
  dim above/.style={to path={\pgfextra{
        \pgfinterruptpath
        \draw[>=latex,|->|] let
        \p1=($(\tikztostart)!1.5em!90:(\tikztotarget)$),
        \p2=($(\tikztotarget)!1.5em!-90:(\tikztostart)$)
        in(\p1) -- (\p2) node[pos=.5,sloped,above]{#1};
        \endpgfinterruptpath
      }
    }
  },
  dim double above/.style={to path={\pgfextra{
        \pgfinterruptpath
        \draw[>=latex,|->|] let
        \p1=($(\tikztostart)!3em!90:(\tikztotarget)$),
        \p2=($(\tikztotarget)!3em!-90:(\tikztostart)$)
        in(\p1) -- (\p2) node[pos=.5,sloped,above]{#1};
        \endpgfinterruptpath
      }
    }
  },
  dim below/.style={to path={\pgfextra{
        \pgfinterruptpath
        \draw[>=latex,|->|] let 
        \p1=($(\tikztostart)!-1em!-90:(\tikztotarget)$),
        \p2=($(\tikztotarget)!-1em!90:(\tikztostart)$)
        in (\p1) -- (\p2) node[pos=.5,sloped,below]{#1};
        \endpgfinterruptpath
      }
    }
  },
}

\tikzset{
    right angle quadrant/.code={
        \pgfmathsetmacro\quadranta{{1,1,-1,-1}[#1-1]}     
        \pgfmathsetmacro\quadrantb{{1,-1,-1,1}[#1-1]}},
    right angle quadrant=1, 
    right angle length/.code={\def\rightanglelength{#1}},   
    right angle length=2ex, 
    right angle symbol/.style n args={3}{
        insert path={
            let \p0 = ($(#1)!(#3)!(#2)$) in     
                let \p1 = ($(\p0)!\quadranta*\rightanglelength!(#3)$), 
                \p2 = ($(\p0)!\quadrantb*\rightanglelength!(#2)$) in 
                let \p3 = ($(\p1)+(\p2)-(\p0)$) in  
            (\p1) -- (\p3) -- (\p2)
        }
    }
}

\newcommand{\pgfextractangle}[3]{%
    \pgfmathanglebetweenpoints{\pgfpointanchor{#2}{center}}
                              {\pgfpointanchor{#3}{center}}
    \global\let#1\pgfmathresult  
}

\usetikzlibrary{shapes.arrows}
\newcommand{\commentarrow}[1][Azure4]{\tikz[baseline=-3pt]{\node[shape border uses incircle, fill=#1,rotate=180,single arrow, inner sep=1pt, minimum size=6pt, single arrow head extend=2pt]{};}}

\tikzset{ax/.style={-latex,line width=2pt}}

\tikzset{camera/.style={fill=Sienna1,fill opacity=0.5},%
image plane/.style={draw=RoyalBlue3,line width=2pt}}

\usepackage{cuted}
\usepackage{caption}
\usepackage[export]{adjustbox}
\usepackage{tabularx}
\usepackage{subcaption}
\usepackage{arydshln}
\title{MPC-Injection: Biasing Off-Policy Locomotion RL Toward Controller-Induced Behavior Basins}

\author{
    Roy Xing$^{1}$, Seyoung Ree$^{2}$, Brian Plancher$^{1}$\\
    $^1$ Dartmouth College,
    $^2$ Harvard University\\
    \texttt{\url{https://a2r-lab.org/MPC-Injection/}}\\
}

\usepackage{placeins}

\begin{document}

\maketitle

\begin{figure}[h]
    \centering
    \includegraphics[
        width=\textwidth,
        clip,
        trim=0cm 0cm 0cm 0.0cm 
        ]{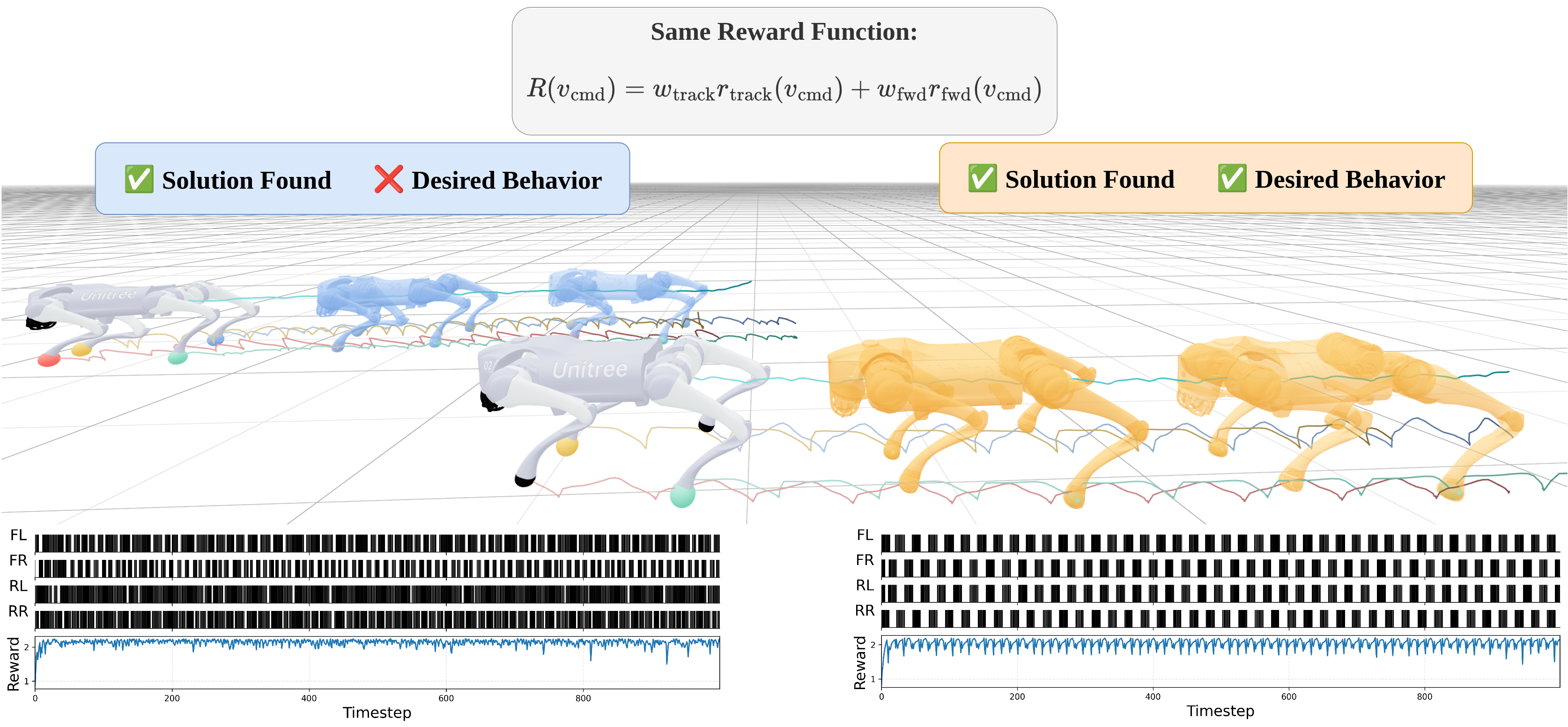}
    \caption{Snapshots of two RL policies trained with the same velocity-tracking reward. Left: pure RL learns to vibrate its legs, producing irregular footsteps. Right: RL with 25\% MPC-Injection learns to trot. Colored spheres mark foot contacts and the raster plots below show footstep timing. Both policies achieve similar return, but only the MPC-Injection policy produces a deployable gait. A more dramatic case on a 2D walker appears in Figure~\ref{fig:walker_diff_policies_same_reward}.}
    \label{fig:diff_policies_same_reward}
\end{figure}

\begin{abstract}
    Reinforcement learning (RL) for locomotion frequently converges to locally optimal but undeployable behaviors, such as vibrating limbs or scooting on the torso, that maximize return without producing a usable gait. We present MPC-Injection, a low-overhead method that steers RL toward a designer-preferred behavior by inserting transitions generated in the same environment by a model predictive controller (MPC). Unlike reward shaping, MPC-Injection does not require redesigning the task reward, and unlike adversarial imitation learning, it adds no discriminator, no kinematic retargeting, and no auxiliary objective. We analyze how the injected transitions bias the learning, allowing the policy to converge to behaviors that pure RL may fail to reach under simple reward functions. On a 2D walker in simulation and with sim-to-real evaluation on a Go2 quadruped, we show that MPC-Injection produces gaits qualitatively comparable to those of reward shaping and adversarial motion priors. We also show that MPC-Injection can complete a barrel roll that pure RL fails to achieve under the same simple reward and can select between trotting and bounding gaits only through changing the injected MPC data.
\end{abstract}


\section{Introduction}
Off-policy reinforcement learning for legged locomotion routinely converges to behaviors that are technically optimal but practically useless. A biped trained with a simple velocity-tracking reward learns to drag itself along the ground because scooting is a stable attractor under that reward. A quadruped trained with a similar reward learns to vibrate some of its limbs, hitting the velocity target while producing a gait that no engineer would deploy on hardware. Both policies achieve high return. Neither produces a usable robot.
This gap between maximizing a reward and producing a deployable behavior is widely recognized. Tasks that admit multiple high-return solutions admit multiple \emph{behavior basins}, regions of the state-visitation space that achieve similar return but differ in kinematics and dynamics. We use \emph{behavior biasing} to mean selecting, among the high-return basins available under a given task reward, the basin preferred by the designer. The central question of this paper is as follows: \emph{how do we effectively and efficiently execute behavior biasing?}

The dominant approaches impose substantial specification costs. Reward shaping adds hand-tuned terms to encourage desired behaviors, requiring extensive trial and error and retraining \cite{muller_olaf_2025,liang_learning_2026,xue_opening_2025}. Imitation approaches, such as adversarial motion priors \cite{peng_amp_2021,li_learning_2022} and trajectory-tracking methods \cite{zhao_resmimic_2025,ze_twist_2025}, reduce reward engineering but require a discriminator network, kinematic retargeting, and a reference motion dataset that must be collected or curated.

We study \emph{MPC-Injection} as a third option. Here a Model Predictive Controller (MPC) generates trajectories in the RL simulation environment, and the resulting transitions are inserted into an off-policy replay buffer. These transitions change the states used in actor updates and the state-action tuples used in critic updates, biasing the learned policy toward the controller's behavior basin. The actor nevertheless selects its own actions rather than imitating recorded MPC actions. The task reward remains unchanged. No behavior-cloning loss, discriminator, or auxiliary objective is introduced, and the policy continues to learn from online experience unlike offline RL.

While similar mechanisms have appeared under different names in prior work (e.g., used to improve return in sparse-reward navigation \cite{dawood_handling_2023, shin_infusing_2022} and to bootstrap manipulation controllers \cite{brudigam_jacta_2024}), to our knowledge, \emph{this is the first analysis of replay-distribution biasing as a mechanism for selecting among distinct high-return behavior basins under an intentionally underspecified locomotion reward}.

Our primary contributions are:
\begin{itemize}[noitemsep, topsep=0pt]
    \item An analysis and empirical characterization of MPC-Injection on a 2D walker, showing that it selects a different behavior basin from vanilla RL under an identical reward, with 25\% injection reliably inducing the controller's basin across SAC and TD3.
    
    \item A Unitree Go2 evaluation showing that MPC-Injection produces gaits qualitatively comparable to twenty-one-term reward shaping and AMP trained on the same MPC data, despite using a two-term task reward. The evaluation includes qualitative sim-to-real transfer and simulation comparisons of torque usage, velocity tracking, and push recovery.

    \item Barrel-roll and bounding experiments extend behavior biasing beyond basic trotting. Here, 25\% MPC-Injection completes a roll that pure RL fails under the same simple reward, and solely replacing trotting data with bounding data selects a bounding gait.
\end{itemize}

\section{Related Work}
Behavior biasing for legged locomotion is typically achieved by either engineering the reward or imitating reference motions. Both approaches transfer the burden of behavior specification onto a particular interface, either the reward function or the motion dataset, and pay an engineering cost proportional to the complexity of the desired behavior. Table~\ref{tab:method_comparison} positions MPC-Injection against these alternatives along three axes: the amount of reward engineering required, the imitation machinery added to the RL loss, and the type of behavior data used.

\begin{table}[t]
  \centering
  \caption{Comparison of MPC-Injection with related behavior biasing methods.}
  \label{tab:method_comparison}
  \small
  \renewcommand{\arraystretch}{1.0}
  \setlength{\tabcolsep}{4pt}
  \begin{tabularx}{\linewidth}{
    >{\raggedright\arraybackslash}p{0.20\linewidth}
    >{\centering\arraybackslash}X
    >{\centering\arraybackslash}X
    >{\centering\arraybackslash}X
  }
    \toprule
    \textbf{Method} &
    \textbf{Reward engineering} &
    \textbf{Imitation machinery} &
    \textbf{Behavior data} \\
    \midrule
    Reward Shaping &
    High: hand-tuned terms &
    None &
    None \\
    \hdashline
    Trajectory tracking IL &
    Low &
    High: tracking loss, retargeting &
    Mocap, reference trajectories \\
    \hdashline
    AMP / adversarial IL &
    Low &
    High: discriminator, retargeting &
    Mocap, motion clips \\
    \hdashline
    Human demo-augmented replay &
    Varies &
    Moderate: replay augmentation &
    Human / expert transitions \\
    \midrule
    \textbf{MPC-Injection (ours)} &
    \textbf{Low} &
    \textbf{None} &
    \textbf{MPC transitions} \\
    \bottomrule
  \end{tabularx}
\end{table}

\textbf{Reward Shaping} \; \;
Reward shaping modifies the task reward by adding hand-tuned terms that reward or penalize specific motions until a desired behavior emerges \cite{laud_theory_2004, ng_policy_1999}. The shaping terms take many forms, including potential functions \cite{jeon_benchmarking_2023, malysheva_learning_2018, harutyunyan_expressing_2015, westenbroek_lyapunov_2022}, barrier functions \cite{kim_learning_2025, kim_not_2024, yang_cbf-rl_2025}, penalty terms \cite{ji_concurrent_2022, rudin_learning_2022}, and phase-modulated terms \cite{siekmann_sim--real_2021, lee_learning_2020, liang_learning_2026}. While effective, reward shaping requires extensive trial-and-error tuning, and any change to the desired behavior typically requires retraining.

\textbf{Imitation Learning} \; \;
Imitation learning sidesteps reward engineering by training the policy to match reference motion data \cite{escontrela_adversarial_2022, pollard_adapting_2002, grimes_dynamic_2006, kurtz_generative_2025}. In legged locomotion, the dominant approaches either track reference trajectories \cite{zhao_resmimic_2025, chen_gmt_2025, fuchioka_opt-mimic_2023} or use adversarial imitation \cite{escontrela_adversarial_2022} as in adversarial motion priors \cite{peng_amp_2021, li_learning_2022}. Both approaches replace reward engineering with separate engineering burdens: kinematic retargeting of motion data to the target robot, and either a tracking loss or a discriminator network that must be trained alongside the policy.

\textbf{Human Demonstration-Augmented Replays} \; \;
Demonstration-augmented off-policy RL uses human expert demonstrations as additional replay data \cite{george_minimizing_2023, vecerik_leveraging_2018, nair_overcoming_2018}. These methods rely on a human demonstrator and typically introduce additional machinery such as prioritized sampling or separate demonstration buffers. The reliance on human teleoperation also limits applicability as collecting demonstrations is impractical for many tasks and embodiments, particularly for morphologies where human intuition does not transfer (see Section~\ref{appendix:robo_morphologies} for a 3-legged half-cheetah example).

\textbf{Hybrid MPC and RL Frameworks} \; \;
A broader literature combines MPC and RL in other ways, including RL residual policies that modify MPC outputs \cite{cheng_rambo_2025, jeon_residual_2025}, MPC as a source of reference trajectories \cite{bogdanovic_model-free_2022}, guided policy search \cite{carius_mpc-net_2020, pmlr-v28-levine13}, and RL-tuned MPC parameters \cite{amos_differentiable_2019, adabag_differentiable_2025}. In the taxonomy of Reiter et al.~\cite{reiter_synthesis_2026}, MPC-Injection falls under MPC for pre-processing in RL training. The methods above either couple the controller to the deployed policy architecturally or use controller outputs as supervision targets, whereas MPC-Injection introduces the controller's behavior only through off-policy transitions and leaves the RL pipeline otherwise unchanged.

\textbf{MPC-Injection} \; \;
Prior work has used controller-generated replay data to bootstrap off-policy RL in sparse-reward navigation \cite{dawood_handling_2023, shin_infusing_2022} and dexterous manipulation \cite{brudigam_jacta_2024, kucuktabak_real-world_2026}, treating injected transitions primarily as a tool for improving return or accelerating exploration. Our work is closest to~\cite{brudigam_jacta_2024} but differs in two ways. First, we show that the mechanism generalizes to a variety of off-policy RL algorithms (specifically SAC and TD3) rather than DDPG alone. Second, we analyze controller-generated replay as a mechanism for selecting among distinct high-return behavior basins under an intentionally underspecified locomotion reward, an analysis that prior work has not provided.

\section{Method}
\begin{figure}[ht]
    \centering
    \includegraphics[clip, trim=0cm 0cm 0cm 0cm, width=0.9\linewidth]{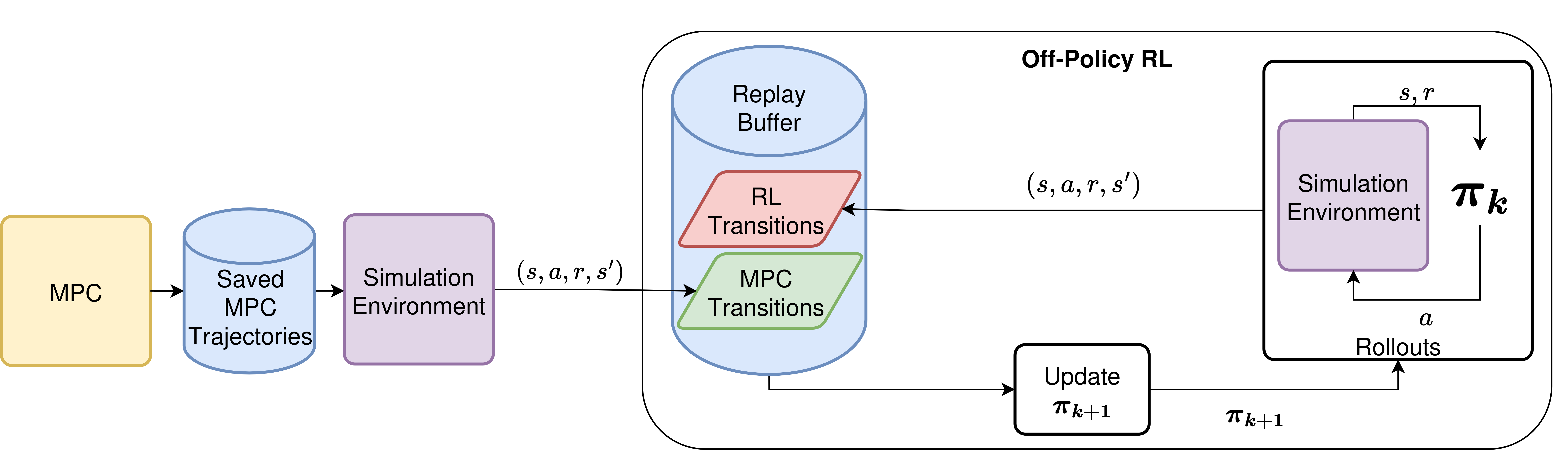}
    \caption{MPC-Injection: An MPC controller is deployed in the same simulation environment used for RL training and the resulting trajectories are stored. During RL training, a target fraction of MPC transitions are injected into the replay buffer alongside on-policy transitions, biasing actor and critic updates toward the controller's behavior basin while leaving the task reward unchanged.}
    \label{fig:mpc-injection}
\end{figure}

MPC-Injection biases the learned policy toward a controller-induced behavior basin by changing the replay state distribution. Injected transitions change both the states at which the actor is optimized and the transition tuples used to train the critic. Actor updates use current-policy actions rather than directly imitating the recorded MPC actions. Figure~\ref{fig:mpc-injection} illustrates the pipeline. The full off-policy training procedure is given as Algorithm~\ref{alg:off_policy_mpc_injection} in Appendix~\ref{appendix:prob_statement}, and Appendix~\ref{appendix:why_mpc_rl_works} provides an optimization-level account of why this induces basin selection rather than behavior cloning.

\textbf{Data Generation} \; \;
We generate MPC trajectories offline in the same simulation environment used for RL training, unlike prior work that generates trajectories online \cite{dawood_handling_2023, shin_infusing_2022}. Offline generation is similar to the pipeline of \cite{brudigam_jacta_2024} and enables precise control over the MPC-to-RL transition ratio in the replay buffer. The MPC objective is designed to produce the desired behavior under the same task, moving the behavior-specification interface from the RL reward into the MPC cost. We use sampling-based MPC \cite{howell2022} for the walker and gradient-based MPC \cite{amatucci_primal-dual_2026} for the quadruped, showing compatibility with both controller classes in the evaluated systems.

Off-the-shelf controller frameworks and solvers reduced the MPC engineering burden in our experiments: controller setup took less than half a day per experiment. For the walker, we also tested sampling- and gradient-based solvers and obtained nearly identical learning results.

Rewards are recomputed from the RL task reward at injection time rather than stored with the MPC trajectories. This allows different RL reward functions to be evaluated against the same MPC dataset without regenerating trajectories.

\textbf{Replay Buffer Injection} \; \;
We define the injection ratio $p$ as the target fraction of replay-buffer entries generated by MPC. For $p<100\%$, after each online transition is inserted, MPC transitions are added until $|\mathcal{D}_\text{MPC}| / |\mathcal{D}|$ reaches $p$; for $p=100\%$, only MPC transitions are stored. Minibatches are sampled uniformly from the combined buffer, so the expected fraction of MPC samples per gradient update is also approximately $p$. MPC transitions enter learning only through the replay distribution: there is no separate imitation loss, no sample weighting, and no auxiliary objective. The training process of the off-policy agent is otherwise unchanged from standard SAC~\cite{haarnoja_soft_2018} or TD3~\cite{fujimoto_addressing_2018}.

\section{Experiments} \label{sec:experiments}
Through our experiments, we evaluate two claims. First, MPC-Injection selects qualitatively different behavior basins from vanilla RL under identical task rewards, with 25\% as the injection ratio that most reliably induces the controller's basin. Second, the resulting behavior is qualitatively similar to reward shaping and AMP trained with MPC data (AMP-MPC) in simulation. Recall that a behavior basin is a region of state space that achieves similar return but may differ in kinematics and dynamics.

Section~\ref{sec:behavior_biasing_under_and_identical_reward} establishes the first claim on a 2D walker~\cite{tassa2018deepmindcontrolsuite} and a Unitree Go2~\cite{unitree_go2} quadruped in simulation. The walker serves as a controlled diagnostic with a deliberately underspecified reward and is evaluated using both SAC and TD3 to confirm that our result is not specific to a single off-policy algorithm. Since the two algorithms produce nearly identical results (Figure~\ref{fig:learning_rate_sac_td3_mpc}), subsequent analysis uses SAC.
Section~\ref{sec:hardware_comparisons} establishes the second claim on the Go2 in simulation and with qualitative sim-to-real evaluation on hardware, demonstrating real-world practicality of MPC-Injection.
Section~\ref{sec:additional_behavior_tasks} evaluates behavior biasing beyond velocity-tracking trotting locomotion through barrel-roll and bounding tasks.
Finally, Section~\ref{appendix:robo_morphologies} provides a qualitative case study on an unconventional morphology, illustrating the data-source advantage of MPC-Injection when reference motions are difficult or impossible to obtain.

\subsection{Experimental Setup}
\textbf{Environments and training} \; \;
We evaluate policies via state-distribution embeddings, footstep regularity, torque usage, and qualitative gait comparison. We use the 2D walker environment from the DeepMind Control Suite~\cite{tassa2018deepmindcontrolsuite} and a Unitree Go2 quadruped, both simulated in MuJoCo~\cite{todorov2012mujoco}. We train SAC and TD3 on the walker for 500,000 environment steps with 5 seeds per configuration, and SAC on the quadruped for 1 million environment steps with 5 seeds. Full network architectures and hyperparameters can be found in Appendix~\ref{appendix:training_details}.

\textbf{Reward functions} \; \;
The walker uses a torso velocity-tracking reward that admits multiple high-return solutions. Here, $v_{\text{torso}}$ denotes the torso velocity and $v_{\text{cmd}}$ denotes the commanded velocity:
\begin{equation}
    R(v_{\text{cmd}}) = (5 \cdot \text{clip}\left (\frac{v_{\text{torso}}}{v_{\text{cmd}}}, 0, 1 \right ) + 1) / 6.
    \label{eq:walker_rwrd_fxn}
\end{equation}
The quadruped uses a two-term reward combining velocity tracking with a forward-progress bias (full definitions in Appendix~\ref{appendix:reward_fxns}, Table~\ref{tab:simple_reward_active}):
\begin{equation}
    R(v_\text{cmd}) = w_\text{track} r_\text{track}(v_\text{cmd}) + w_\text{fwd} r_\text{fwd} (v_\text{cmd}).
    \label{eq:high_lvl_quadruped_rwrd_fxn_mpc_injection}
\end{equation}
The reward-shaping baseline, adapted from the Unitree mjlab implementation \cite{unitree_rl_mjlab} with term weights re-tuned for off-policy training, uses twenty-one tuned terms. The evaluated AMP-MPC baseline uses an off-the-shelf AMP-PPO implementation trained from the same underlying Go2 MPC rollouts as MPC-Injection (Tables~\ref{tab:go2_amp_reward_composition}, \ref{tab:go2_amp_reward_positive}, and \ref{tab:go2_amp_reward_penalty}).

\textbf{MPC trajectory generation and sim-to-real} \; \;
We use sampling-based MPC \cite{howell2022} for the walker and gradient-based MPC \cite{amatucci_primal-dual_2026} for the quadruped, with trajectories generated offline in the same simulation environment used for RL training. For the quadruped, the MPC outputs joint torques while the RL policy outputs target joint position offsets. For sim-to-real transfer on the Go2, we apply domain randomization over mass, friction, and joint dynamics (Appendix~\ref{appendix:sim2real}).


\subsection{Replay Buffer Distribution Biasing Under an Identical Reward} \label{sec:behavior_biasing_under_and_identical_reward}

We evaluate the claim that MPC-Injection selects qualitatively different locomotion basins from vanilla RL under identical task rewards, by sweeping MPC-Injection ratios on the 2D walker as a controlled diagnostic, then verifying the same pattern on the Unitree Go2 quadruped.

\begin{figure}[t]
    \centering
    \subfloat[SAC.\label{fig:learning_rate_sac_mpc}]{
        \includegraphics[width=0.48\linewidth]{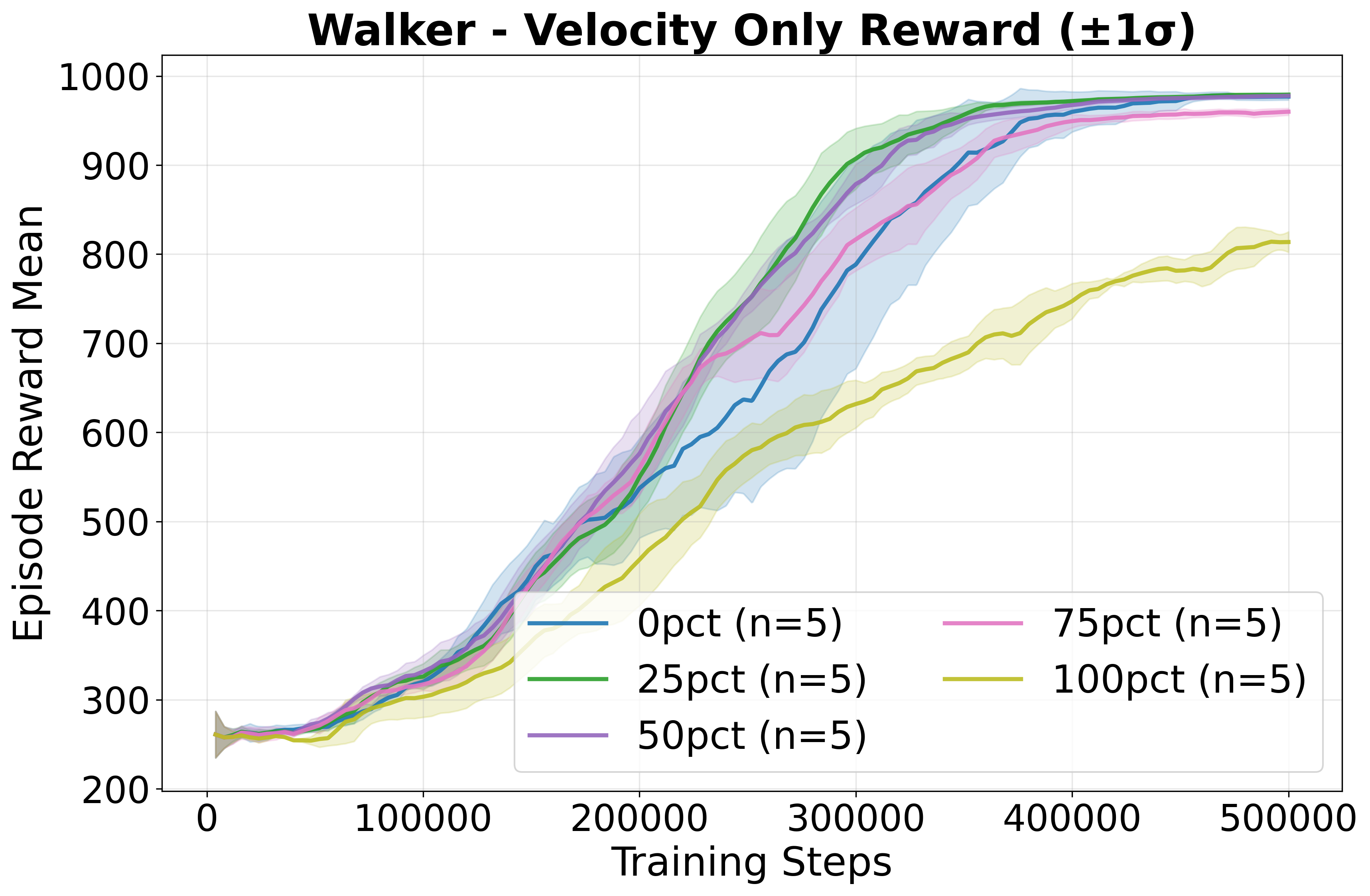}
    }
    \hfill
    \subfloat[TD3.\label{fig:learning_rate_td3_mpc}]{
        \includegraphics[width=0.48\linewidth]{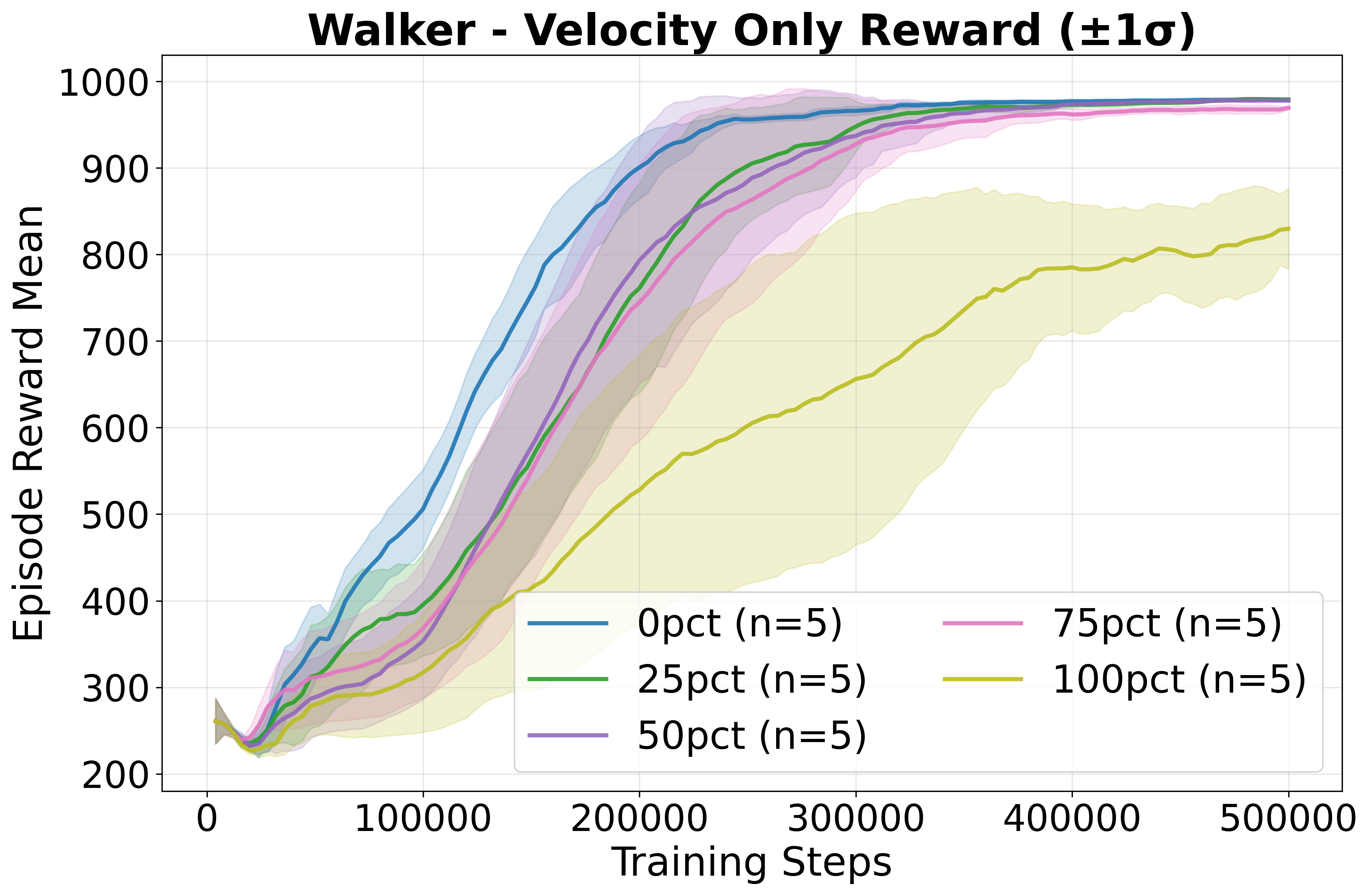}
    }
    \caption{Episodic return on the walker velocity-tracking reward across MPC-Injection ratios (0\%, 25\%, 50\%, 75\%, 100\%) for (a) SAC and (b) TD3. Several injection ratios converge to similar episodic return, confirming that the task reward admits multiple high-return solutions. Because the reward is underspecified for the desired behavior, training curves do not reflect which behavior basin each policy has converged to. Basin membership is evaluated separately in Figures~\ref{fig:umap_state_distr} and~\ref{fig:walker_diff_policies_same_reward}.}
    \label{fig:learning_rate_sac_td3_mpc}
\end{figure}

\textbf{Walker} \; \;
The walker reward in Eq.~\ref{eq:walker_rwrd_fxn} depends only on torso velocity and therefore admits multiple high-return solutions. We sweep injection ratios across 0\% (pure RL), 25\%, 50\%, 75\%, and 100\% on both SAC and TD3. Each ratio was evaluated with five seeds; at the final checkpoint, the controller-like upright gait occurred in all five 25\%-injection runs for both SAC and TD3.
Figure~\ref{fig:learning_rate_sac_td3_mpc} shows that several injection settings reach similar episodic return across both algorithms, confirming that the replay-distribution bias can change behavior without requiring a different task reward. Because the underspecified reward makes return uninformative about basin membership, we evaluate basin membership through qualitative footstep analysis and state-distribution embeddings.

\begin{figure*}[t]
  \centering
  \begin{subfigure}[c]{0.325\textwidth}
      \centering
      \includegraphics[width=\linewidth]{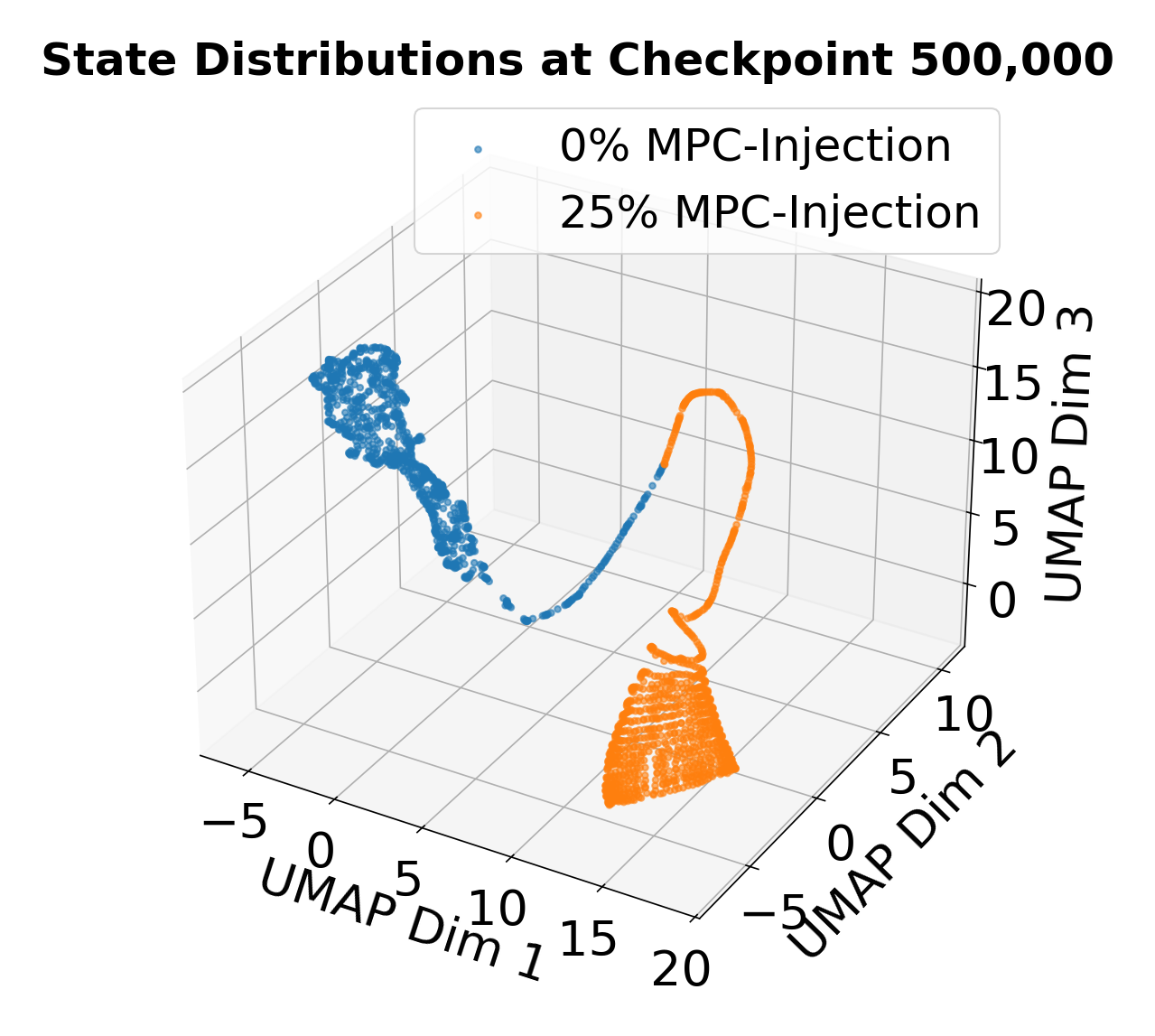}
      \caption{}
      \label{fig:umap_state_distr}
  \end{subfigure}
  \hfill
  \begin{subfigure}[c]{0.65\textwidth}
      \centering
      \includegraphics[width=\linewidth]{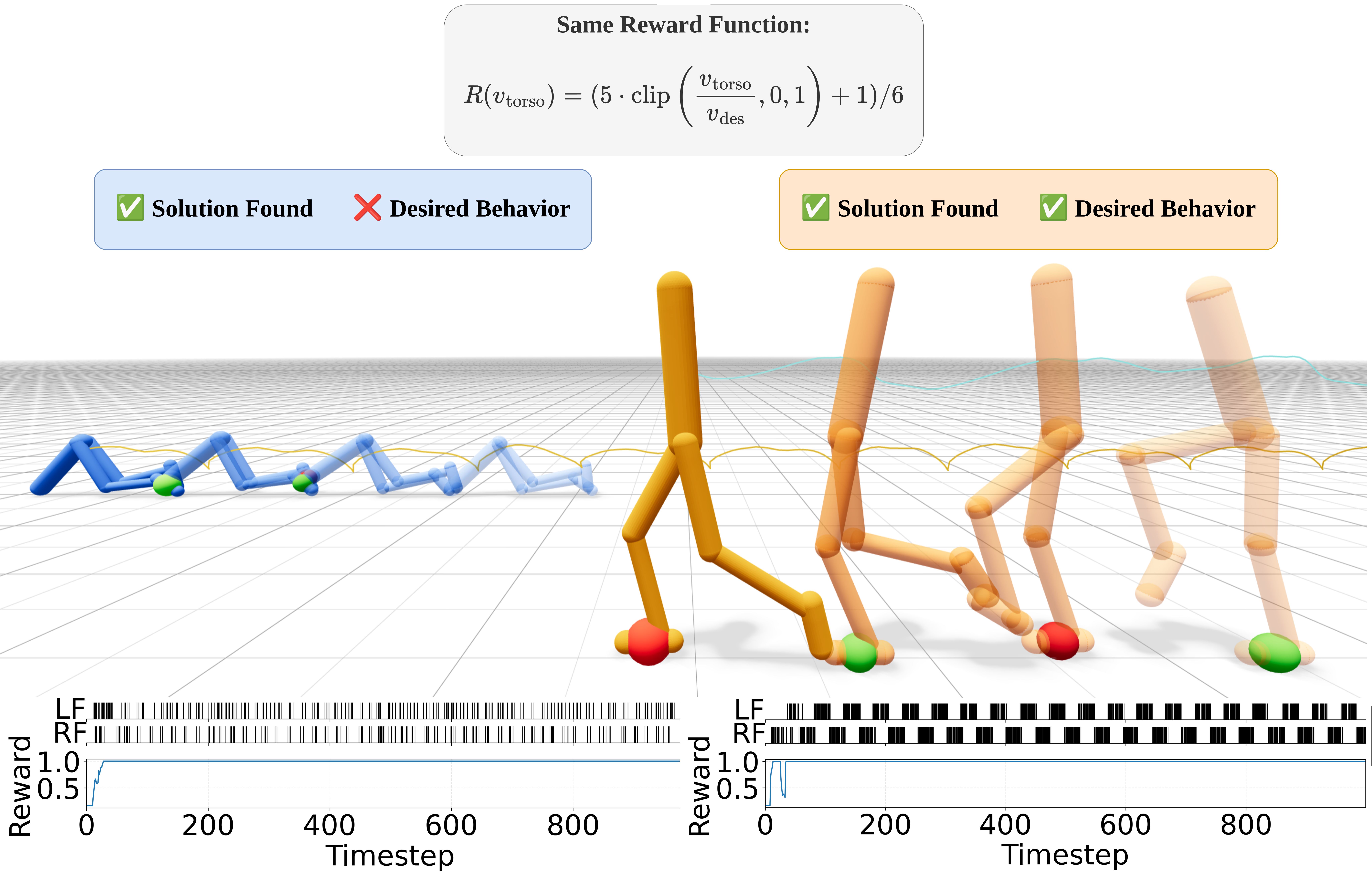}
      \caption{}
      \label{fig:walker_diff_policies_same_reward}
  \end{subfigure}
    \caption{ State-space structure and qualitative behavior under 0\% and 25\% MPC-Injection with the same velocity-tracking reward. (a) UMAP embedding of the visited state distributions showing that 25\% MPC-Injection occupies a compact, structured region of state space while 0\% (pure RL) occupies a more diffuse region. (b) Behavior snapshots and footstep raster plots. The 0\% policy (blue walker, orange torso trajectory) drags its torso along the ground in a scooting gait. The 25\% policy (orange walker, cyan torso trajectory) walks upright with periodic alternating footsteps. Red spheres mark right-foot contacts, green spheres mark left-foot contacts.}
  \label{fig:walker_training_dynamics}
\end{figure*}

Figure~\ref{fig:walker_diff_policies_same_reward} shows that pure RL converges to a grounded scooting strategy that drags the top of the torso along the ground, while 25\% MPC-Injection learns an upright, periodic gait matching the MPC behavior. Figure~\ref{fig:umap_state_distr} corroborates this through the state-distribution embeddings, revealing that the 25\% policy occupies a compact, structured region of state space while the pure RL policy occupies a more diffuse region. Higher injection ratios (50\%, 75\%, 100\%) fail to produce regular gaits, as the actor gradient becomes dominated by states the policy cannot consistently reach via its own rollouts (Appendices~\ref{appendix:why_mpc_rl_works} and~\ref{appendix:mpc-injection_pct_ablation} contain the higher-ratio behaviors and the optimization-level account).
Prior MPC-Injection work \cite{brudigam_jacta_2024, shin_infusing_2022} reports that around 25\% is an ideal injection ratio and improves returned reward in non-locomotion tasks. We find that in locomotion the same injection ratio still works well, but instead selects the controller's behavior basin without necessarily improving return.

\begin{figure}[t]
    \centering
    \subfloat[Pure RL\label{fig:pure_rl_footstep_trajs}]{
        \includegraphics[width=0.425\linewidth]{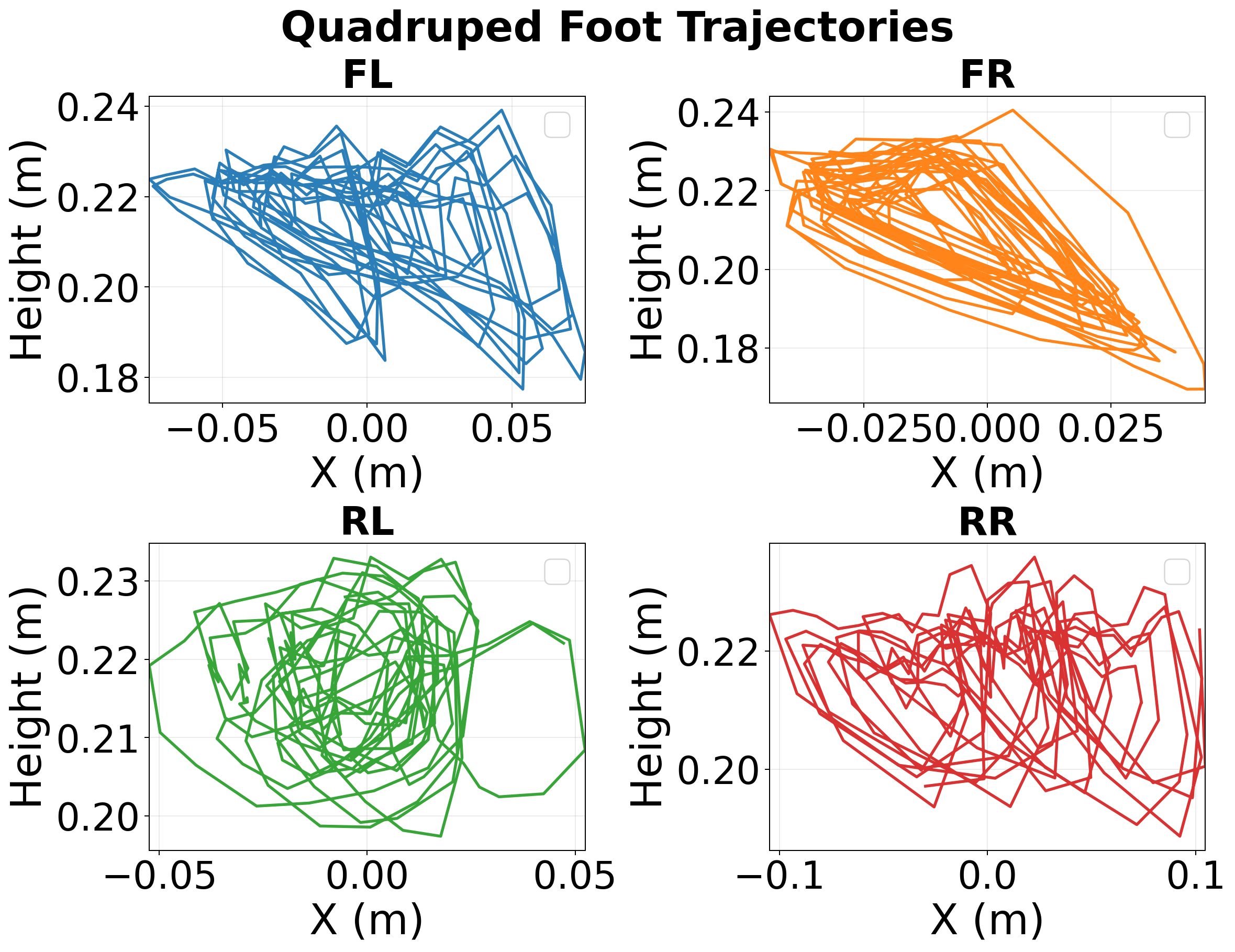}
    }%
    \subfloat[25\% MPC-Injection\label{fig:25pct_footstep_trajs}]{
        \includegraphics[width=0.425\linewidth]{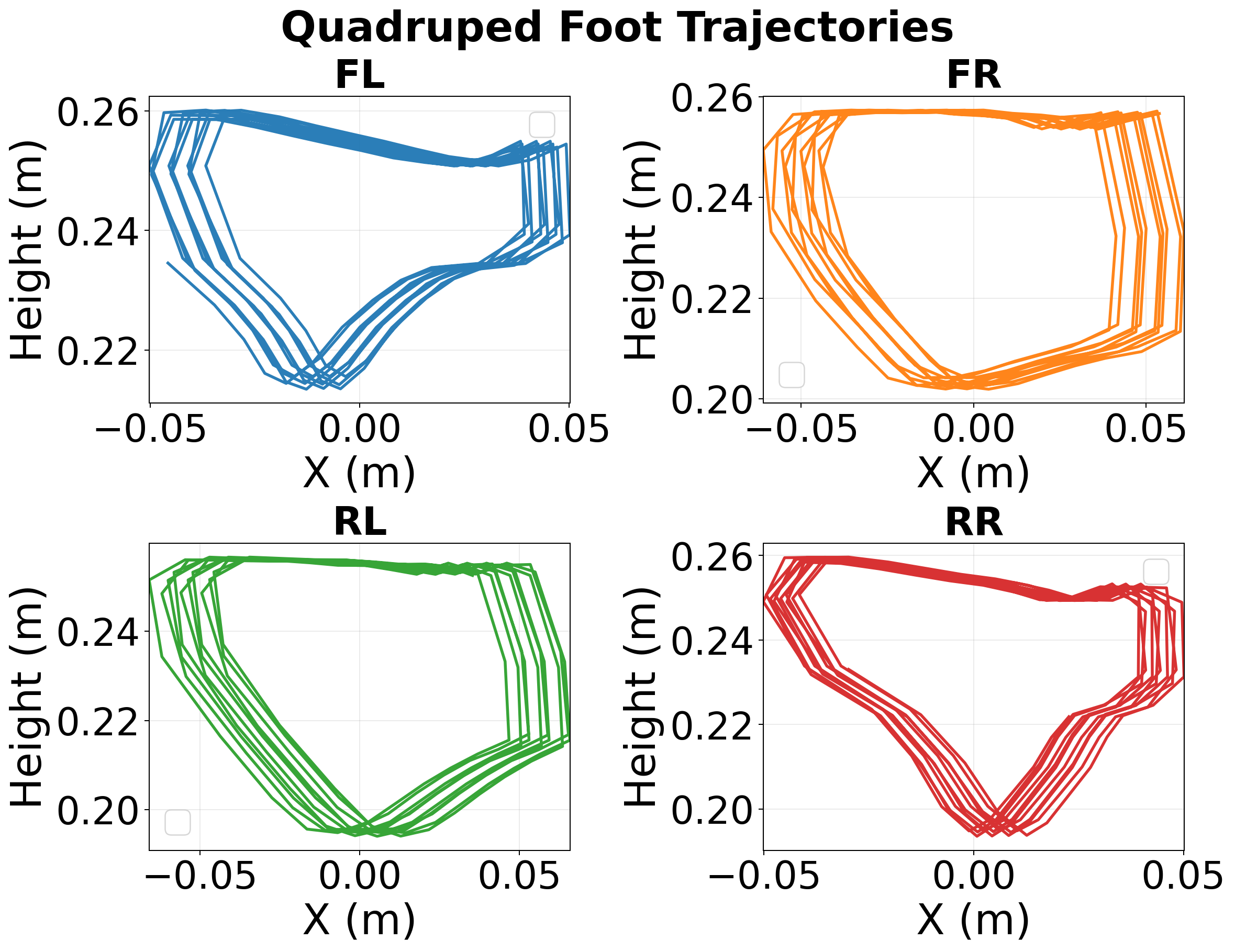}
    }
    \caption{Quadruped footstep comparisons. (a) Footstep trajectories under pure RL show chaotic, irregular patterns. (b) Footstep trajectories under 25\% MPC-Injection show structured, periodic patterns matching a trotting gait.}
    \label{fig:footstep_trajs}
\end{figure}

\begin{figure}[!t]
    \centering
    \begin{subfigure}[b]{0.3\linewidth}
        \centering
        \includegraphics[width=\linewidth]{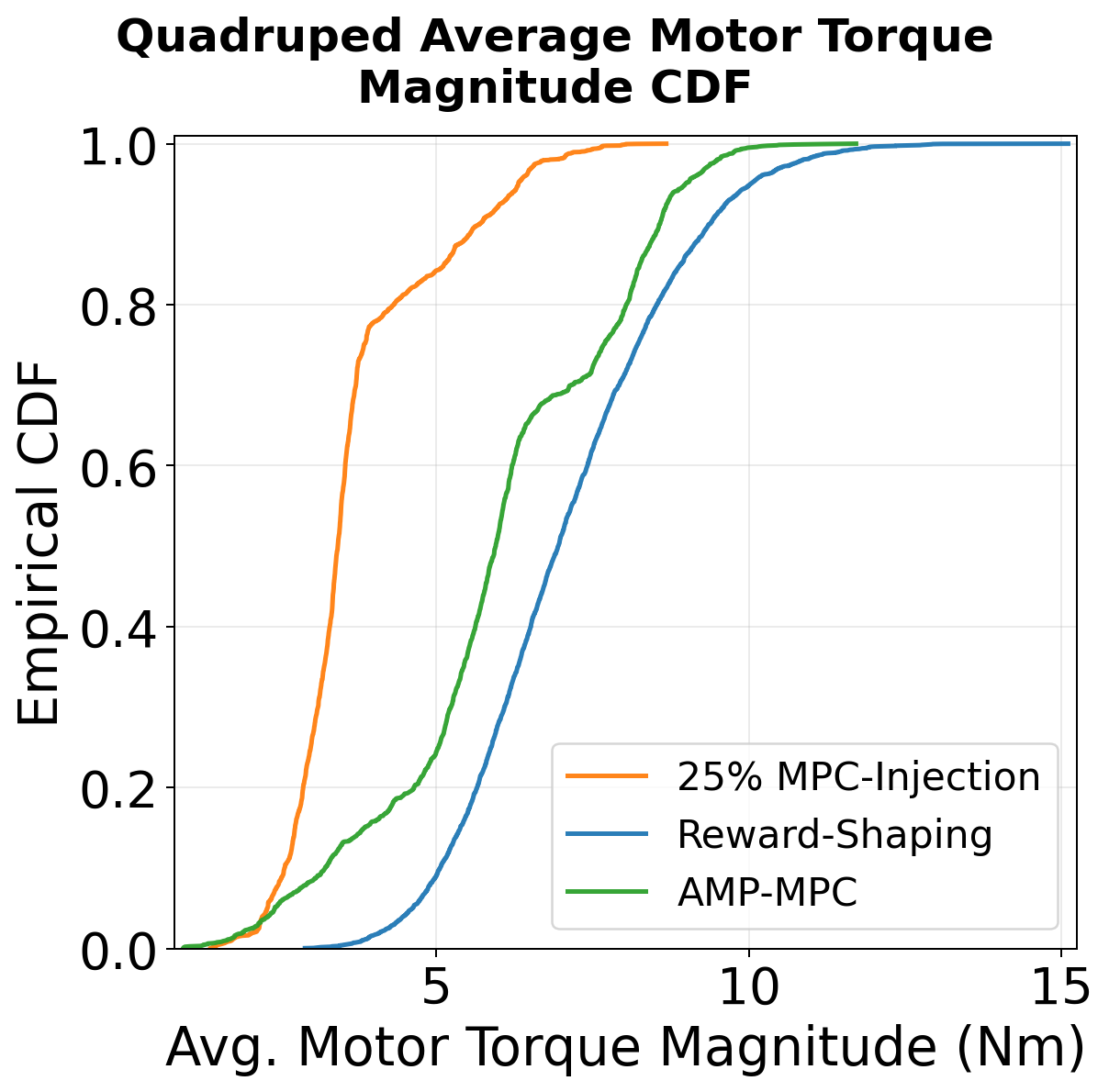}
        \caption{Torque usage CDF}
        \label{fig:torque_usage}
    \end{subfigure}
    \begin{subfigure}[b]{0.3\linewidth}
        \centering
        \includegraphics[width=\linewidth]{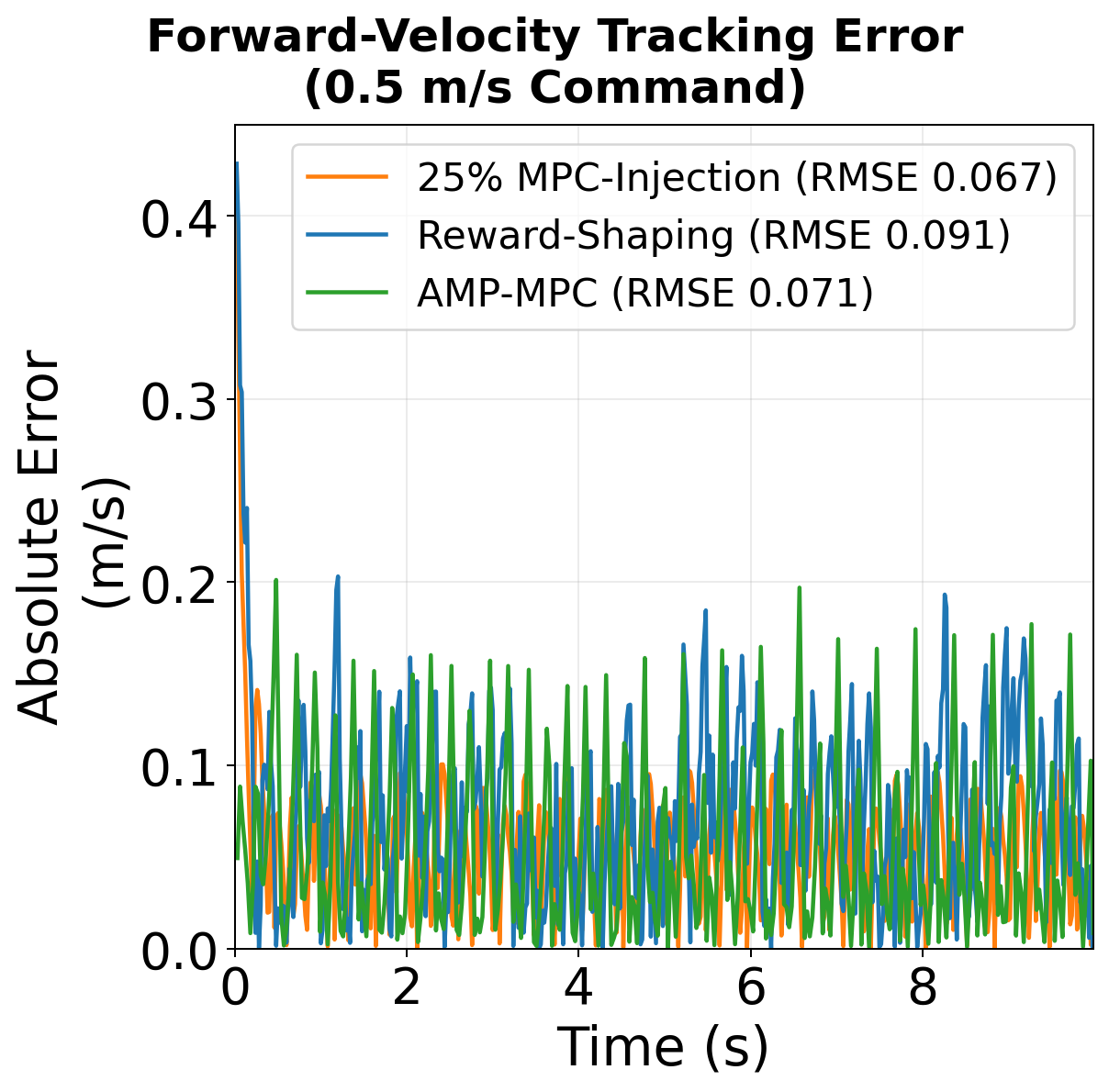}
        \caption{Forward-velocity tracking}
        \label{fig:amp_mpc_tracking_simulation}
    \end{subfigure}
    \caption{Quadruped simulation baseline metrics. (a) The motor-torque magnitude CDF compares 25\% MPC-Injection with reward shaping and AMP-MPC. MPC-Injection uses lower average torque magnitudes. (b) Forward-velocity tracking at a 0.5~m/s command for 25\% MPC-Injection, AMP-MPC, and reward shaping.}
    \label{fig:quadruped_simulation_baseline_metrics}
\end{figure}

\textbf{Quadruped} \; \;
The quadruped exhibits the same qualitative pattern as the walker. Pure off-policy RL learns to vibrate the joints (Figure~\ref{fig:diff_policies_same_reward}), producing chaotic and irregular footstep trajectories (Figure~\ref{fig:pure_rl_footstep_trajs}). With 25\% MPC-Injection, the policy learns a structured trot with periodic footstep patterns (Figure~\ref{fig:25pct_footstep_trajs}). Across five independent Go2 training seeds, the 25\% setting produced the structured controller-like trot in all five final-policy rollouts. An analogous injection-ratio sweep on the quadruped (Appendix~\ref{appendix:mpc-injection_pct_ablation}) confirms that 25\% remains the most reliable operating regime. The torque magnitude CDF in Figure~\ref{fig:torque_usage} further shows lower average motor-torque magnitudes for MPC-Injection than for AMP-MPC and reward shaping, consistent with the energy-cost terms in the MPC objective. Together with the walker results, this confirms that 25\% MPC-Injection consistently induces the controller's behavior basin across both the diagnostic walker and the deployment-target quadruped under simple velocity-tracking rewards.

\begin{table}[!t]
\centering
\caption{Simulated lateral-push success rates across 10 trials.}
\label{tab:lateral_push_robustness}
\small
\begin{tabular}{llccc}
\toprule
& & \multicolumn{3}{c}{Push velocity (m/s)} \\
\cmidrule(lr){3-5}
Policy & Push source & 0.5 & 1.0 & 1.5 \\
\midrule
AMP-MPC
& Left  & 100\% & 100\% & 10\% \\
& Right & 100\% & 50\%  & 0\% \\
\addlinespace
MPC-Inj
& Left  & 100\% & 50\%  & 0\% \\
& Right & 100\% & 70\%  & 0\% \\
\bottomrule
\end{tabular}
\end{table}

\begin{figure}[!t]
    \centering
    \subfloat[Reward-Shaped \label{fig:real_quad_reward_shaped}]{
        \includegraphics[width=0.32\linewidth]{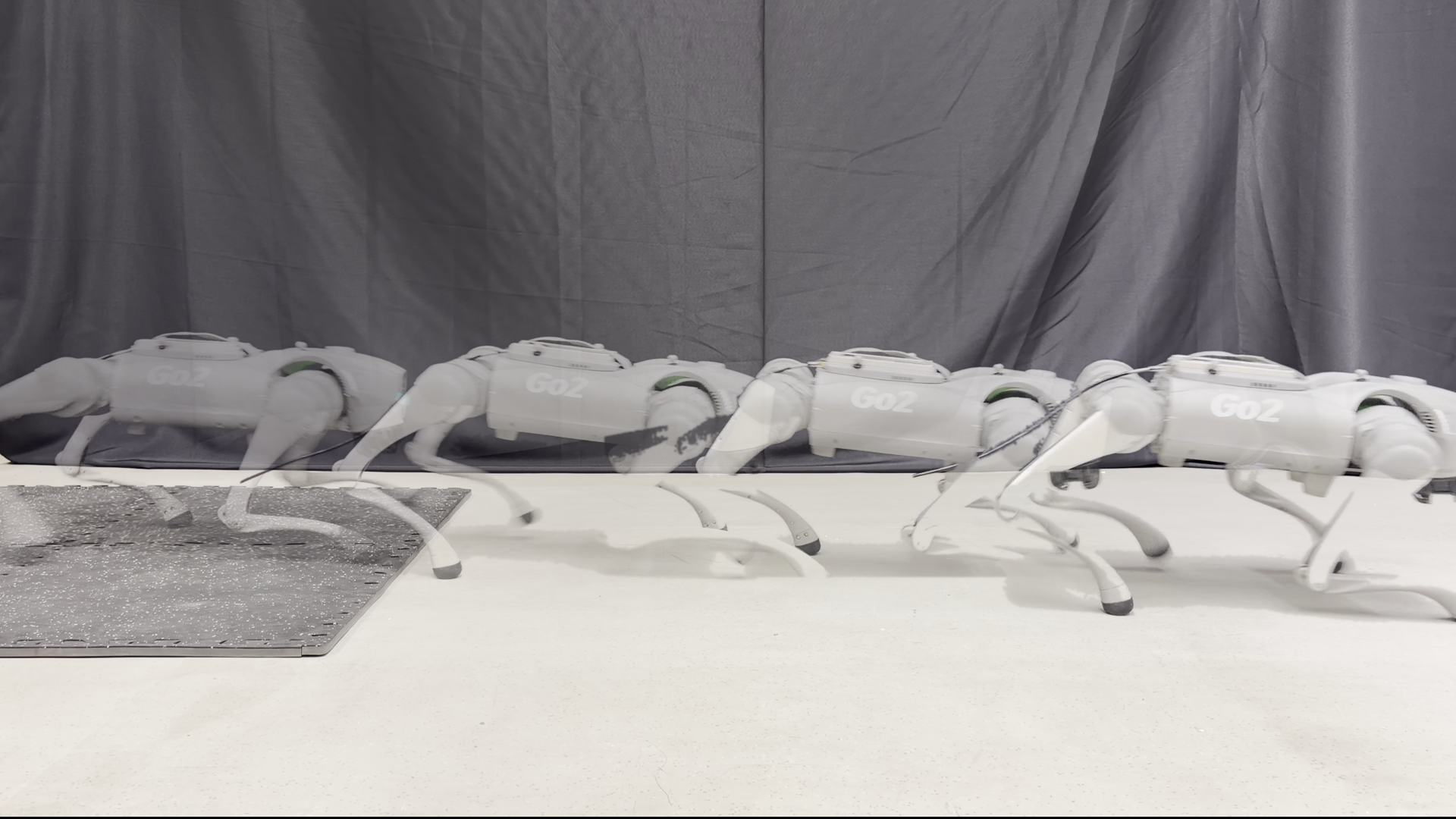}
    }%
    \hfill
    \subfloat[AMP-MPC \label{fig:real_quad_amp}]{
        \includegraphics[width=0.32\linewidth]{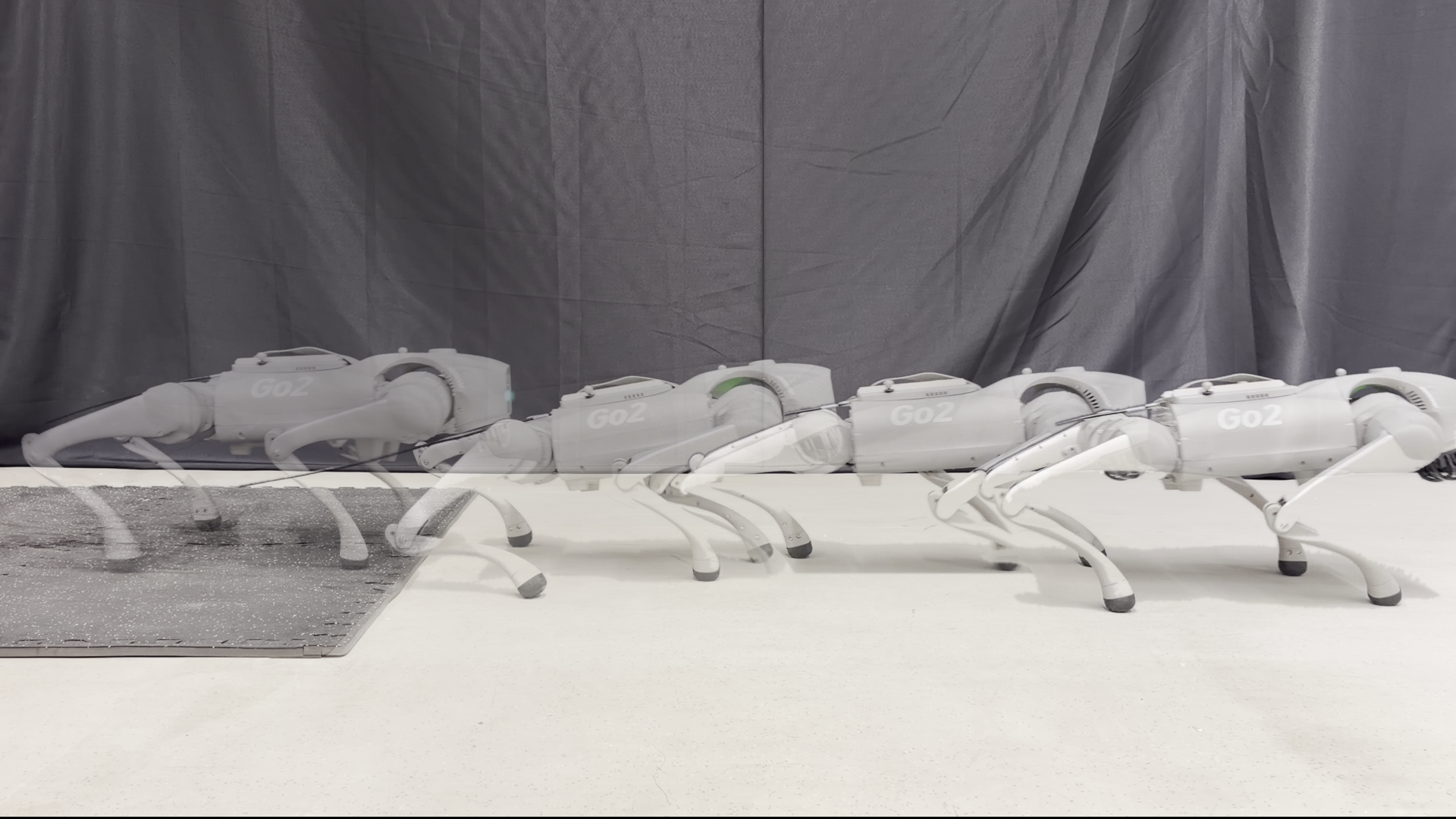}
    }%
    \hfill
    \subfloat[MPC-Injection (ours) \label{fig:real_quad_mpc-injection}]{
        \includegraphics[width=0.32\linewidth]{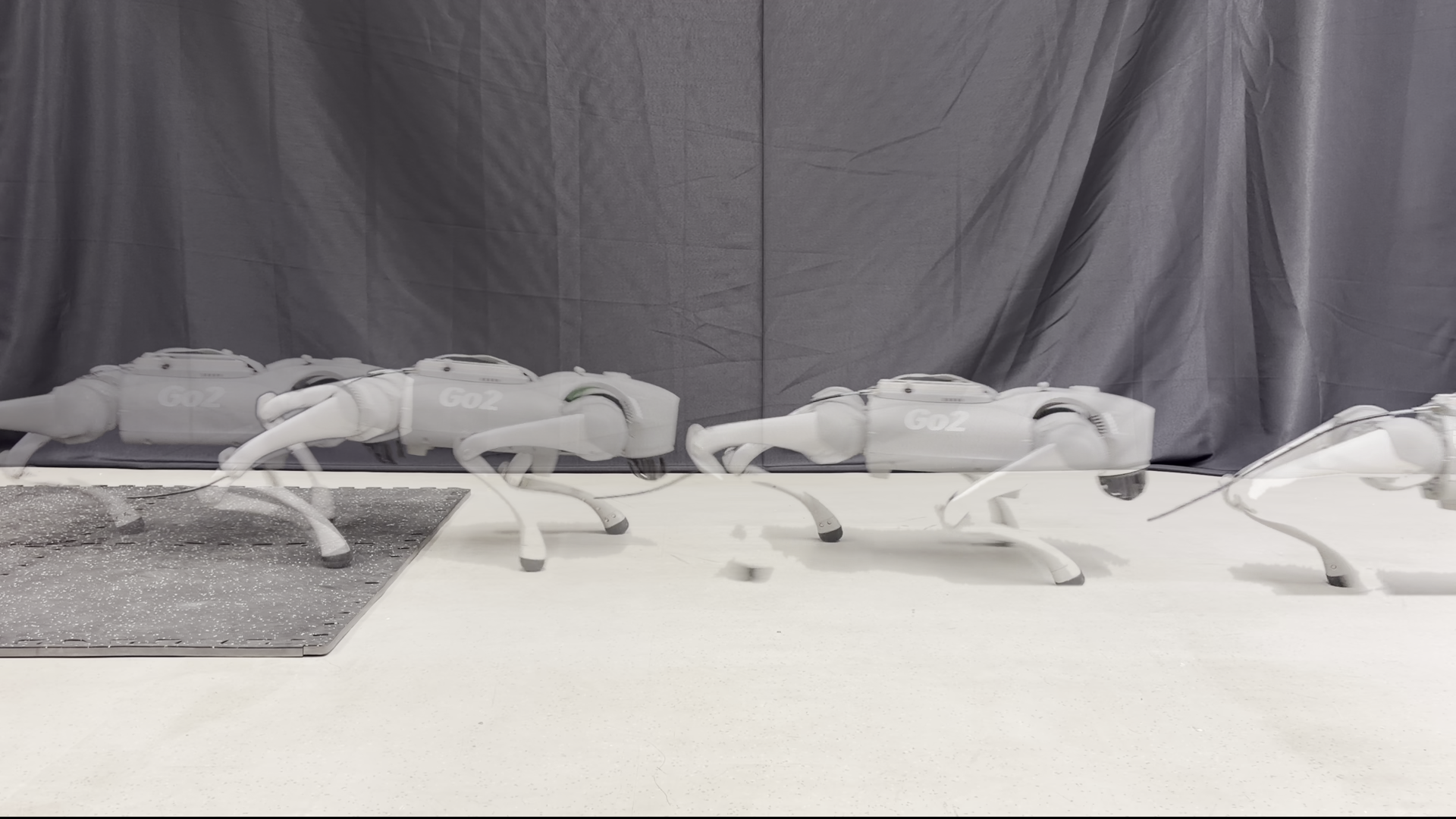}
    }
    \caption{Qualitative snapshots from the reported reward-shaping, AMP-MPC, and MPC-Injection hardware trials.}
    \label{fig:qual_comparisons}
\end{figure}

\subsection{Simulation Comparisons to AMP-MPC and Qualitative Hardware Trials}
\label{sec:hardware_comparisons}
We use an off-the-shelf AMP-PPO implementation with the same asymmetric 45/48 actor--critic observation split used by MPC-Injection. AMP-MPC uses the same underlying MPC data, low-pass filter, PD gains, and domain-randomization setup as MPC-Injection. AMP-MPC produces behavior visually similar to MPC-Injection. Figure~\ref{fig:torque_usage} shows that the 25\% MPC-Injection policy has lower average motor-torque magnitudes than AMP-MPC and reward shaping. At a 0.5~m/s command, Figure~\ref{fig:amp_mpc_tracking_simulation} reports forward-velocity RMSEs of 0.067 for 25\% MPC-Injection, 0.071 for AMP-MPC, and 0.091 for reward shaping.

Table~\ref{tab:lateral_push_robustness} reports simulated lateral-push recovery across ten trials. AMP-MPC has higher success for left pushes at 1.0 and 1.5~m/s, whereas MPC-Injection has higher success for right pushes at 1.0~m/s.

Figure~\ref{fig:qual_comparisons} and the supplemental video on our website show qualitative hardware trials. The reward-shaped policy produces a stable gait with relatively hard foot strikes, while MPC-Injection produces stable trots similar to the simulated MPC controller. AMP-MPC struggles to cross the sim-to-real gap, leading it to shuffle its feet along the ground on real hardware.

\begin{figure}[t]
    \centering
    \begin{minipage}[c]{0.52\linewidth}
        \centering
        \begin{subfigure}[b]{0.48\linewidth}
            \centering
            \includegraphics[width=\linewidth]{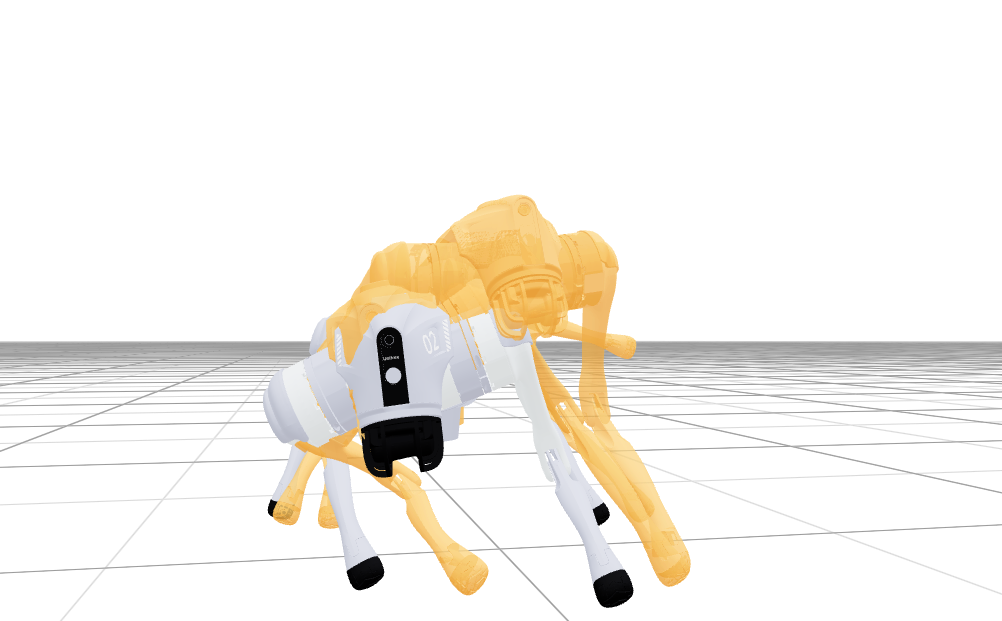}
            \caption{Pure RL barrel roll}
            \label{fig:barrel-roll-pure-rl}
        \end{subfigure}%
        \hspace{0.02\linewidth}
        \begin{subfigure}[b]{0.48\linewidth}
            \centering
            \includegraphics[width=\linewidth]{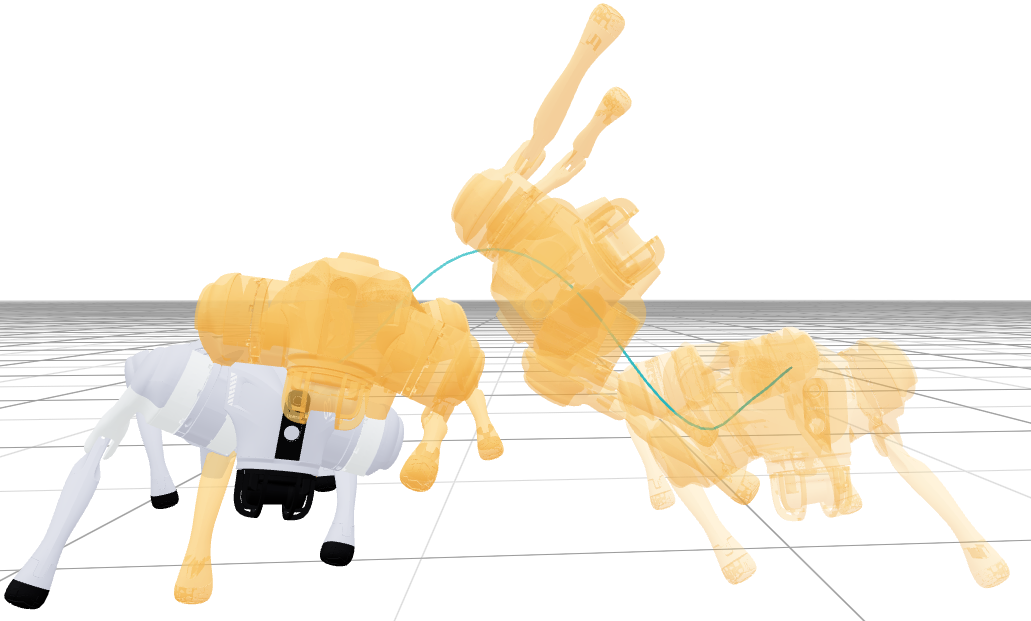}
            \caption{MPC-Inj. barrel roll}
            \label{fig:barrel-roll-mpc-injection}
        \end{subfigure}

        \begin{subfigure}[b]{0.8\linewidth}
            \centering
            \includegraphics[width=\linewidth]{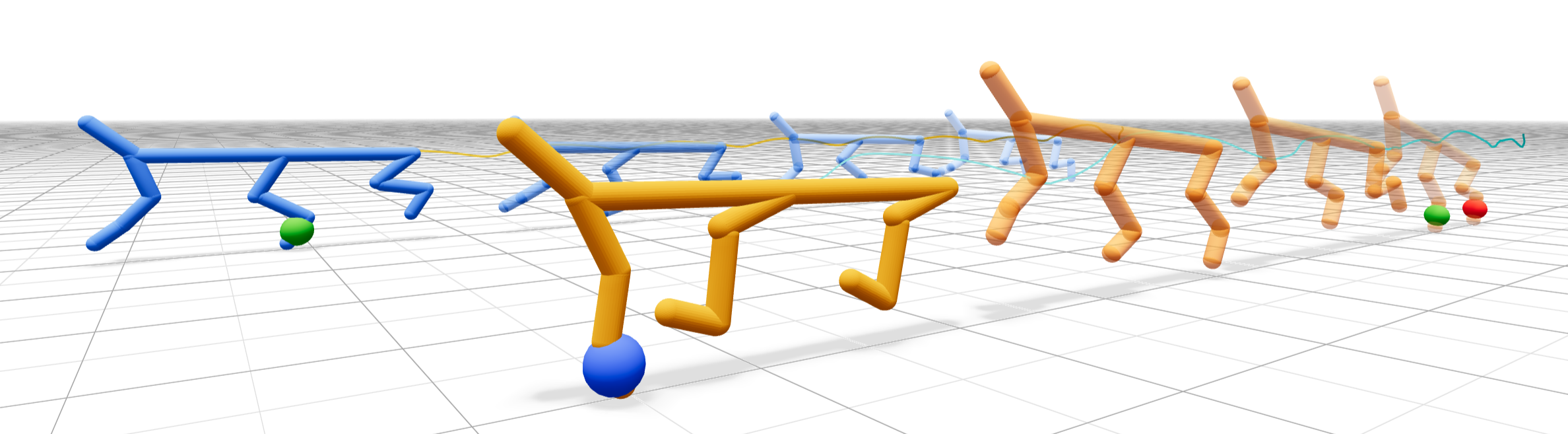}
            \caption{Unusual morphology}
            \label{fig:cheetah3_results}
        \end{subfigure}
    \end{minipage}%
    \hfill
    \begin{minipage}[c]{0.46\linewidth}
        \centering
        \begin{subfigure}[b]{0.85\linewidth}
            \centering
            \includegraphics[width=\linewidth]{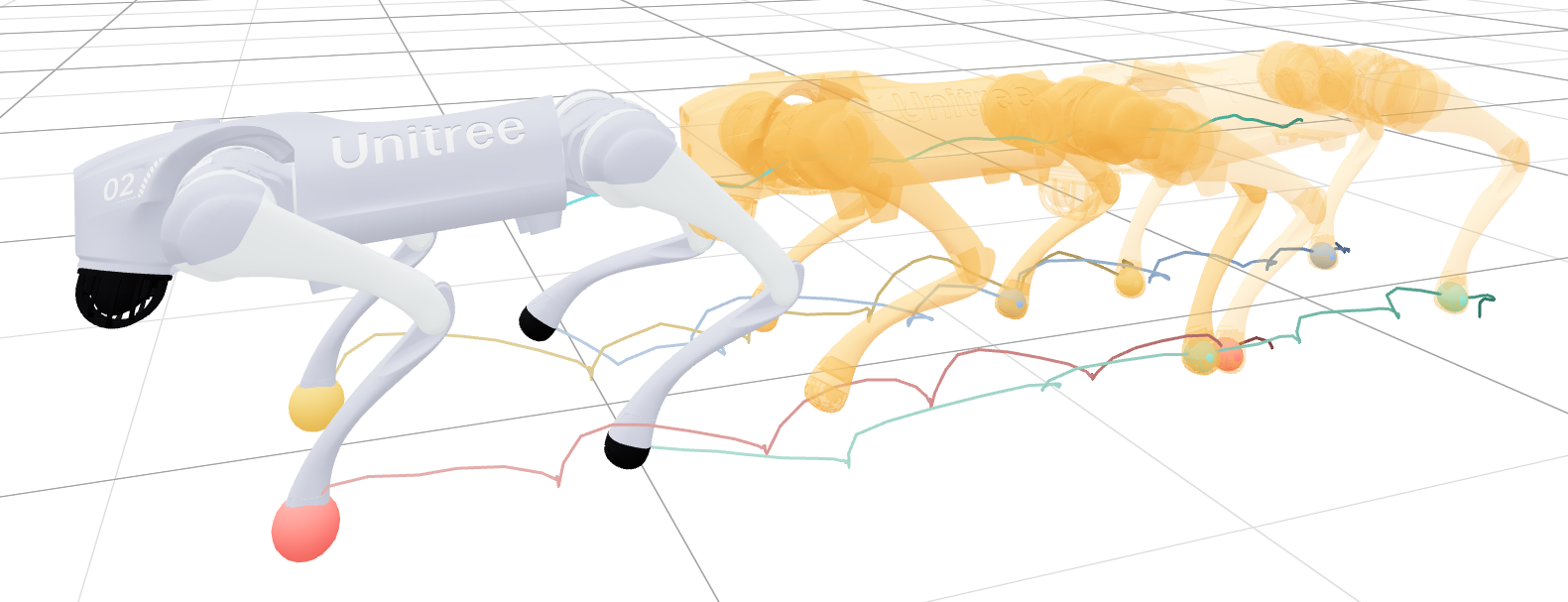}
            \caption{Bounding in simulation}
            \label{fig:bounding-sim}
            \label{fig:bounding}
        \end{subfigure}

        \begin{subfigure}[b]{0.9\linewidth}
            \centering
            \includegraphics[width=\linewidth]{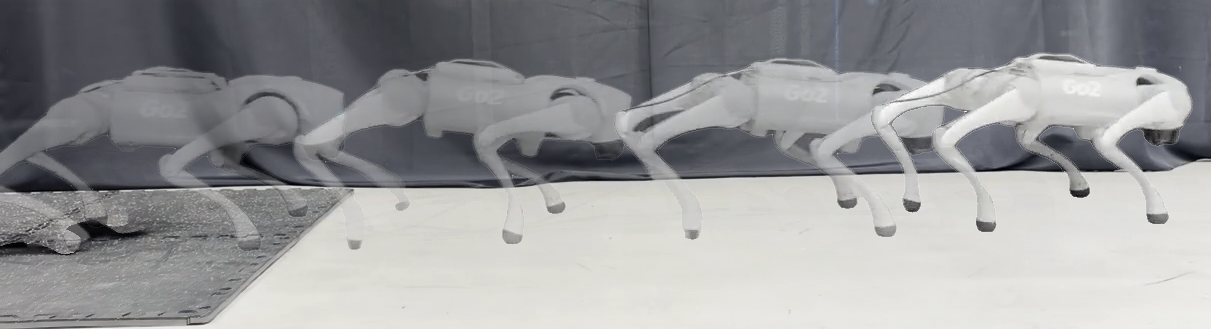}
            \caption{Bounding on hardware}
            \label{fig:bounding-hardware}
        \end{subfigure}
    \end{minipage}
    \caption{Behavior-biasing case studies. MPC-Injection completes the barrel roll (b) while pure RL fails (a). A bounding gait is learned via MPC-Injection (d, e) by changing only the injected MPC data. MPC-Injection also allows for behavior biasing of unusual morphologies as seen in (c).}
    \label{fig:barrel-roll}
\end{figure}

\subsection{Additional Behavior Tasks: Barrel Roll and Bounding} \label{sec:additional_behavior_tasks}
We evaluate a Go2 barrel-roll task to test MPC-Injection beyond velocity-tracking locomotion. Figure~\ref{fig:barrel-roll} shows that pure RL does not complete the barrel roll, whereas 25\% MPC-Injection succeeds under the same simple reward, using the same off-the-shelf MPX controller used for the locomotion task to collect MPC data. 

We note that such behaviors commonly require curriculum learning or assistive wrench forces to help the robot flip, which we can avoid with MPC-Injection \cite{sleiman_zest_2026}. Here, $\tilde{\phi}_t$ and $\phi_t^*$ are the measured and desired unwrapped roll angles, $\Delta\tilde{\phi}_t=\tilde{\phi}_t-\tilde{\phi}_{t-1}$ is the per-step roll progress, $s\in\{-1,1\}$ is roll direction, and $\Delta\phi_{\mathrm{exp}}$ normalizes progress. $\alpha$ controls tracking sensitivity, $c$ bounds rewards, and $r_t^{\mathrm{term}}$ provides terminal success or failure:
\begin{equation*}
r_t =
e^{-\alpha(\phi_t^*-\tilde{\phi}_t)^2}
+\mathrm{clip}\!\left(
\frac{s\,\Delta\tilde{\phi}_t}{\Delta\phi_{\mathrm{exp}}},-c,c
\right)
+r_t^{\mathrm{term}}.
\end{equation*}

Furthermore, only replacing the injected trotting MPC data with bounding MPC data, while leaving the task reward and training procedure unchanged, induces a bounding gait (Figure~\ref{fig:bounding}). Thus, changing only the injected controller data can select a different behavior basin, reinforcing the power and simplicity of our core mechanism.

\subsection{Qualitative Case Study on MPC-Injection for Unusual Robot Morphologies} \label{appendix:robo_morphologies}

Finally, we illustrate that MPC can be particularly useful as a behavior-data source for non-standard locomotion tasks, especially those with no reference dataset and no human teleoperation demonstrations. In these settings, adversarial imitation methods and trajectory-tracking methods cannot be directly applied without first synthesizing reference motions through a separate process. MPC, by contrast, can be deployed on any morphology for which a dynamics model and cost function are available. As such, we include a qualitative case study on an unconventional morphology: a 3-legged half-cheetah modified from the DM Control Suite.
Figure~\ref{fig:cheetah3_results} compares pure RL and 25\% MPC-Injection under the same simple torso-velocity-tracking reward for this unique morphology. As with our prior experiments, the pure RL policy converges to an undesirable strategy, two-legged walking that leaves the front leg vestigial. In contrast, the 25\% MPC-Injection policy uses all three legs in a coordinated gait that mirrors the optimized MPC behavior.

\FloatBarrier

\section{Conclusion}

We presented MPC-Injection, a low-overhead method for behavior biasing in off-policy RL that selects a designer-preferred locomotion behavior by inserting controller-generated transitions into the replay buffer. On a 2D walker and a Unitree Go2 quadruped, MPC-Injection drives the learned policy into the controller's behavior basin using only a one- to two-term task reward. It produces locomotion qualitatively comparable to reward shaping with twenty-one tuned terms and to AMP-MPC trained on the same controller data, while using lower average motor-torque magnitudes in simulation and transferring qualitatively to Go2 hardware. Go2 barrel-roll and bounding tasks further demonstrate behavior biasing beyond velocity-tracking trotting locomotion, including the selection of a different gait by changing only the injected MPC data. These results position MPC-Injection as a practical alternative to reward shaping and imitation learning when a controller for the desired behavior already exists, or is easy to design.

\section{Limitations \& Future Work}

Several limitations bound our results. First, the optimization-level analysis in Appendix~\ref{appendix:why_mpc_rl_works} predicts that the implicit basin pull from replay-distribution biasing may weaken under extended training. Once the policy operates inside the controller's basin, on-policy rollouts overlap heavily with injected MPC states, reducing the additional pull contributed by those transitions. Preliminary experiments on the quadruped are consistent with this prediction, but not conclusive. Future work could study mechanisms for preserving basin membership during late training, such as curriculum-based injection, basin-aware early stopping, or critics augmented with behavior-sensitive regularization. 
Second, the current paper studies behavior biasing primarily in relatively simple velocity-tracking locomotion tasks, with additional barrel-roll and bounding case studies. It remains to be shown how robust the method is under broader command spaces, more diverse terrains, contact-rich tasks, and online or adaptive MPC data generation. In particular, we found that naively mixing trajectories generated from widely varying velocity commands does not always preserve a clean behavior basin during learning. This motivates future work on curriculum learning over command ranges and online selection of which MPC trajectories to inject.
Finally, MPC-Injection should not be interpreted as eliminating behavior specification but as shifting the burden from reward tuning or imitation-learning pipeline development to standard controller design, and is therefore most useful when a controller for the desired behavior already exists.

\acknowledgments{
We used LLM tools, including Codex and Claude Code, to assist with software implementation, proofreading, and review. All final text and references were edited and reviewed by humans.
This work is supported by the National Science Foundation under Grants DGE-2236868 and 2411369. Any opinions, findings, conclusions, or recommendations expressed in this material are those of the authors and may not reflect those of the NSF.
The authors would also like to thank Andrea Grillo for his help during filming and perturbation testing.
}

\FloatBarrier
\bibliography{biblio/IEEEFull,biblio/IEEEConfFull,biblio/OtherFull,
biblio/mpc_for_pre_post_processing, 
biblio/reward_shaping,
biblio/imitation_learning,
biblio/gradient_mpc,
biblio/sampling_mpc,
biblio/model_free_rl,
biblio/residual_policies,
biblio/closed_loop_learning_mpc_rl,
biblio/replay_buff_from_demos}

@string{ARXIV = "ar{X}iv"}

@string{ICML = "International Conference on Machine Learning"}

@inproceedings{amos_differentiable_2019,
	title = {Differentiable {MPC} for {End}-to-end {Planning} and {Control}},
	url = {https://proceedings.neurips.cc/paper_files/paper/2018/hash/ba6d843eb4251a4526ce65d1807a9309-Abstract.html},
	language = {en},
	author = {Amos, Brandon and Jimenez, Ivan and Sacks, Jacob and Boots, Byron and Kolter, J. Zico},
	year = {2018},
	booktitle = {Advances in Neural Information Processing Systems},
	volume = {31},
	pages = {8289--8300},
	publisher = {Curran Associates, Inc.},
}

@inproceedings{adabag_differentiable_2025,
	title = {Differentiable {Model} {Predictive} {Control} on the {GPU}},
	url = {https://iclr.cc/virtual/2026/oral/10008631},
	language = {en},
	author = {Adabag, Emre and Greiff, Marcus and Subosits, John and Lew, Thomas},
	year = {2026},
	booktitle = {International Conference on Learning Representations},
}

@article{kim_contact-implicit_2025,
  title   = {Contact-Implicit Model Predictive Control: Controlling Diverse Quadruped Motions Without Pre-Planned Contact Modes or Trajectories},
  author  = {Kim, Gijeong and Kang, Dongyun and Kim, Joon-Ha and Hong, Seungwoo and Park, Hae-Won},
  journal = {The International Journal of Robotics Research},
  volume  = {44},
  number  = {3},
  pages   = {486--510},
  year    = {2025},
  month   = mar,
  doi     = {10.1177/02783649241273645},
  url     = {https://doi.org/10.1177/02783649241273645}
}

@inproceedings{du_gato_2025,
  title     = {GATO: GPU-Accelerated and Batched Trajectory Optimization for Scalable Edge Model Predictive Control},
  author    = {Du, Alexander and Adabag, Emre and Bravo Palacios, Gabriel and Plancher, Brian},
  booktitle = {2026 IEEE International Conference on Robotics and Automation (ICRA)},
  year      = {2026}
}

@inproceedings{alavilli_tinympc_2025,
  title     = {TinyMPC: Model-Predictive Control on Resource-Constrained Microcontrollers},
  author    = {Nguyen, Khai and Schoedel, Sam and Alavilli, Anoushka and Plancher, Brian and Manchester, Zachary},
  booktitle = {2024 IEEE International Conference on Robotics and Automation (ICRA)},
  year      = {2024},
  doi       = {10.1109/ICRA57147.2024.10610987},
  url       = {https://doi.org/10.1109/ICRA57147.2024.10610987}
}

@article{peng_amp_2021,
	title = {{AMP}: {Adversarial} {Motion} {Priors} for {Stylized} {Physics}-{Based} {Character} {Control}},
	volume = {40},
	issn = {0730-0301, 1557-7368},
	shorttitle = {{AMP}},
	url = {http://arxiv.org/abs/2104.02180},
	doi = {10.1145/3450626.3459670},
	language = {en},
	number = {4},
	urldate = {2025-10-10},
	journal = {ACM Transactions on Graphics},
	author = {Peng, Xue Bin and Ma, Ze and Abbeel, Pieter and Levine, Sergey and Kanazawa, Angjoo},
	month = aug,
	year = {2021},
	note = {arXiv:2104.02180 [cs]},
	pages = {1--20},
}

@inproceedings{pollard_adapting_2002,
	title = {Adapting human motion for the control of a humanoid robot},
	volume = {2},
	url = {https://ieeexplore.ieee.org/document/1014737/},
	doi = {10.1109/ROBOT.2002.1014737},
	urldate = {2025-12-10},
	booktitle = {Proceedings 2002 {IEEE} {International} {Conference} on {Robotics} and {Automation} ({Cat}. {No}.{02CH37292})},
	author = {Pollard, N.S. and Hodgins, J.K. and Riley, M.J. and Atkeson, C.G.},
	month = may,
	year = {2002},
	pages = {1390--1397 vol.2},
}

@inproceedings{grimes_dynamic_2006,
	title = {Dynamic {Imitation} in a {Humanoid} {Robot} through {Nonparametric} {Probabilistic} {Inference}},
	isbn = {978-0-262-69348-6},
	url = {http://www.roboticsproceedings.org/rss02/p26.pdf},
	doi = {10.15607/RSS.2006.II.026},
	language = {en},
	urldate = {2025-12-10},
	booktitle = {Robotics: {Science} and {Systems} {II}},
	publisher = {Robotics: Science and Systems Foundation},
	author = {Grimes, D. and Chalodhorn, R. and Rao, R.},
	month = aug,
	year = {2006},
}

@misc{zhao_resmimic_2025,
	title = {{ResMimic}: {From} {General} {Motion} {Tracking} to {Humanoid} {Whole}-body {Loco}-{Manipulation} via {Residual} {Learning}},
	shorttitle = {{ResMimic}},
	url = {http://arxiv.org/abs/2510.05070},
	doi = {10.48550/arXiv.2510.05070},
	language = {en},
	urldate = {2025-12-10},
	publisher = {arXiv},
	author = {Zhao, Siheng and Ze, Yanjie and Wang, Yue and Liu, C. Karen and Abbeel, Pieter and Shi, Guanya and Duan, Rocky},
	month = oct,
	year = {2025},
	note = {arXiv:2510.05070 [cs]},
}

@misc{chen_gmt_2025,
	title = {{GMT}: {General} {Motion} {Tracking} for {Humanoid} {Whole}-{Body} {Control}},
	shorttitle = {{GMT}},
	url = {http://arxiv.org/abs/2506.14770},
	doi = {10.48550/arXiv.2506.14770},
	language = {en},
	urldate = {2025-12-10},
	publisher = {arXiv},
	author = {Chen, Zixuan and Ji, Mazeyu and Cheng, Xuxin and Peng, Xuanbin and Peng, Xue Bin and Wang, Xiaolong},
	month = sep,
	year = {2025},
	note = {arXiv:2506.14770 [cs]},
}

@inproceedings{li_learning_2022,
	title = {Learning {Agile} {Skills} via {Adversarial} {Imitation} of {Rough} {Partial} {Demonstrations}},
	url = {https://proceedings.mlr.press/v205/li23b.html},
	language = {en},
	publisher = {PMLR},
	author = {Li, Chenhao and Vlastelica, Marin and Blaes, Sebastian and Frey, Jonas and Grimminger, Felix and Martius, Georg},
	year = {2023},
	booktitle = {Proceedings of the 6th Conference on Robot Learning},
	series = {Proceedings of Machine Learning Research},
	volume = {205},
	pages = {342--352},
}

@inproceedings{ze_twist_2025,
	title = {{TWIST}: {Teleoperated} {Whole}-{Body} {Imitation} {System}},
	shorttitle = {{TWIST}},
	url = {https://proceedings.mlr.press/v305/ze25a.html},
	language = {en},
	publisher = {PMLR},
	author = {Ze, Yanjie and Chen, Zixuan and Araújo, João Pedro and Cao, Zi-ang and Peng, Xue Bin and Wu, Jiajun and Liu, C. Karen},
	year = {2025},
	booktitle = {Proceedings of the 9th Conference on Robot Learning},
	series = {Proceedings of Machine Learning Research},
	volume = {305},
	pages = {2143--2154},
}

@inproceedings{fuchioka_opt-mimic_2023,
	address = {London, United Kingdom},
	title = {{OPT}-{Mimic}: {Imitation} of {Optimized} {Trajectories} for {Dynamic} {Quadruped} {Behaviors}},
	copyright = {https://doi.org/10.15223/policy-029},
	isbn = {979-8-3503-2365-8},
	shorttitle = {{OPT}-{Mimic}},
	url = {https://ieeexplore.ieee.org/document/10160562/},
	doi = {10.1109/ICRA48891.2023.10160562},
	urldate = {2026-04-02},
	booktitle = {2023 {IEEE} {International} {Conference} on {Robotics} and {Automation} ({ICRA})},
	publisher = {IEEE},
	author = {Fuchioka, Yuni and Xie, Zhaoming and Van De Panne, Michiel},
	month = may,
	year = {2023},
	pages = {5092--5098},
}

@inproceedings{escontrela_adversarial_2022,
	address = {Kyoto, Japan},
	title = {Adversarial {Motion} {Priors} {Make} {Good} {Substitutes} for {Complex} {Reward} {Functions}},
	copyright = {https://doi.org/10.15223/policy-029},
	isbn = {978-1-6654-7927-1},
	url = {https://ieeexplore.ieee.org/document/9981973/},
	doi = {10.1109/IROS47612.2022.9981973},
	urldate = {2026-04-02},
	booktitle = {2022 {IEEE}/{RSJ} {International} {Conference} on {Intelligent} {Robots} and {Systems} ({IROS})},
	publisher = {IEEE},
	author = {Escontrela, Alejandro and Peng, Xue Bin and Yu, Wenhao and Zhang, Tingnan and Iscen, Atil and Goldberg, Ken and Abbeel, Pieter},
	month = oct,
	year = {2022},
	pages = {25--32},
}

@article{stable-baselines3,
  author  = {Antonin Raffin and Ashley Hill and Adam Gleave and Anssi Kanervisto and Maximilian Ernestus and Noah Dormann},
  title   = {Stable-Baselines3: Reliable Reinforcement Learning Implementations},
  journal = {Journal of Machine Learning Research},
  year    = {2021},
  volume  = {22},
  number  = {268},
  pages   = {1-8},
  url     = {http://jmlr.org/papers/v22/20-1364.html}
}

@inproceedings{todorov2012mujoco,
  title={MuJoCo: A physics engine for model-based control},
  author={Todorov, Emanuel and Erez, Tom and Tassa, Yuval},
  booktitle={2012 IEEE/RSJ International Conference on Intelligent Robots and Systems},
  pages={5026--5033},
  year={2012},
  organization={IEEE},
  doi={10.1109/IROS.2012.6386109}
}

@misc{tassa2018deepmindcontrolsuite,
      title={DeepMind Control Suite}, 
      author={Yuval Tassa and Yotam Doron and Alistair Muldal and Tom Erez and Yazhe Li and Diego de Las Casas and David Budden and Abbas Abdolmaleki and Josh Merel and Andrew Lefrancq and Timothy Lillicrap and Martin Riedmiller},
      year={2018},
      eprint={1801.00690},
      archivePrefix={arXiv},
      primaryClass={cs.AI},
      url={https://arxiv.org/abs/1801.00690}, 
}

@inproceedings{fujimoto_addressing_2018,
  title     = {Addressing Function Approximation Error in Actor-Critic Methods},
  author    = {Fujimoto, Scott and van Hoof, Herke and Meger, David},
  booktitle = {Proceedings of the 35th International Conference on Machine Learning},
  editor    = {Dy, Jennifer and Krause, Andreas},
  series    = {Proceedings of Machine Learning Research},
  volume    = {80},
  pages     = {1587--1596},
  year      = {2018},
  publisher = {PMLR},
  url       = {https://proceedings.mlr.press/v80/fujimoto18a.html}
}

@inproceedings{haarnoja_soft_2018,
  title     = {Soft Actor-Critic: Off-Policy Maximum Entropy Deep Reinforcement Learning with a Stochastic Actor},
  author    = {Haarnoja, Tuomas and Zhou, Aurick and Abbeel, Pieter and Levine, Sergey},
  booktitle = {Proceedings of the 35th International Conference on Machine Learning},
  editor    = {Dy, Jennifer and Krause, Andreas},
  series    = {Proceedings of Machine Learning Research},
  volume    = {80},
  pages     = {1861--1870},
  year      = {2018},
  publisher = {PMLR},
  url       = {https://proceedings.mlr.press/v80/haarnoja18b.html}
}

@article{sleiman_zest_2026,
  title   = {{ZEST}: {Zero}-shot embodied skill transfer for athletic robot control},
  author  = {Sleiman, Jean Pierre and Li, He and Adu-Bredu, Alphonsus and Deits, Robin and Kumar, Arun and Bergamin, Kevin and Bhardwaj, Mohak and Biddlestone, Scott and Burger, Nicola and Estrada, Matthew A. and Iacobelli, Francesco and Koolen, Twan and Lambert, Alexander and Lin, Erica and Mungai, M. Eva and Nobles, Zach and Rozen-Levy, Shane and Shi, Yuyao and Wang, Jiashun and Welner, Jakob and Yu, Fangzhou and Zhang, Mike and Rizzi, Alfred and Hodgins, Jessica and Bertrand, Sylvain and Abe, Yeuhi and Kuindersma, Scott and Farshidian, Farbod},
  journal = {Science Robotics},
  volume  = {11},
  number  = {117},
  pages   = {eaec7695},
  month   = aug,
  year    = {2026},
  doi     = {10.1126/scirobotics.aec7695},
  url     = {https://doi.org/10.1126/scirobotics.aec7695}
}

@inproceedings{brudigam_jacta_2024,
	title = {Jacta: {A} {Versatile} {Planner} for {Learning} {Dexterous} and {Whole}-body {Manipulation}},
	shorttitle = {Jacta},
	url = {https://proceedings.mlr.press/v270/bruedigam25a.html},
	language = {en},
	publisher = {PMLR},
	author = {Brüdigam, Jan and Abbas, Ali-Adeeb and Sorokin, Maks and Fang, Kuan and Hung, Brandon and Guru, Maya and Sosnowski, Stefan and Wang, Jiuguang and Hirche, Sandra and Cleac'h, Simon Le},
	year = {2025},
	booktitle = {Proceedings of the 8th Conference on Robot Learning},
	series = {Proceedings of Machine Learning Research},
	volume = {270},
	pages = {994--1020},
}

@inproceedings{kim_learning_2025,
	address = {Atlanta, GA, USA},
	title = {A {Learning} {Framework} for {Diverse} {Legged} {Robot} {Locomotion} {Using} {Barrier}-{Based} {Style} {Rewards}},
	copyright = {https://doi.org/10.15223/policy-029},
	isbn = {979-8-3315-4139-2},
	url = {https://ieeexplore.ieee.org/document/11128517/},
	doi = {10.1109/ICRA55743.2025.11128517},
	language = {en},
	urldate = {2025-10-28},
	booktitle = {2025 {IEEE} {International} {Conference} on {Robotics} and {Automation} ({ICRA})},
	publisher = {IEEE},
	author = {Kim, Gijeong and Lee, Yong-Hoon and Park, Hae-Won},
	month = may,
	year = {2025},
	pages = {10004--10010},
}

@article{carius_mpc-net_2020,
	title = {{MPC}-{Net}: {A} {First} {Principles} {Guided} {Policy} {Search}},
	volume = {5},
	issn = {2377-3766, 2377-3774},
	shorttitle = {{MPC}-{Net}},
	url = {http://arxiv.org/abs/1909.05197},
	doi = {10.1109/LRA.2020.2974653},
	language = {en},
	number = {2},
	urldate = {2025-11-22},
	journal = {IEEE Robotics and Automation Letters},
	author = {Carius, Jan and Farshidian, Farbod and Hutter, Marco},
	month = apr,
	year = {2020},
	note = {arXiv:1909.05197 [cs]},
	pages = {2897--2904},
}

@misc{vecerik_leveraging_2018,
	title = {Leveraging {Demonstrations} for {Deep} {Reinforcement} {Learning} on {Robotics} {Problems} with {Sparse} {Rewards}},
	url = {http://arxiv.org/abs/1707.08817},
	doi = {10.48550/arXiv.1707.08817},
	language = {en},
	urldate = {2025-12-01},
	publisher = {arXiv},
	author = {Vecerik, Mel and Hester, Todd and Scholz, Jonathan and Wang, Fumin and Pietquin, Olivier and Piot, Bilal and Heess, Nicolas and Rothörl, Thomas and Lampe, Thomas and Riedmiller, Martin},
	month = oct,
	year = {2018},
	note = {arXiv:1707.08817 [cs]},
}

@article{shin_infusing_2022,
	title = {Infusing model predictive control into meta-reinforcement learning for mobile robots in dynamic environments},
	volume = {7},
	issn = {2377-3766, 2377-3774},
	url = {http://arxiv.org/abs/2109.07120},
	doi = {10.1109/LRA.2022.3191234},
	language = {en},
	number = {4},
	urldate = {2025-12-02},
	journal = {IEEE Robotics and Automation Letters},
	author = {Shin, Jaeuk and Hakobyan, Astghik and Park, Mingyu and Kim, Yeoneung and Kim, Gihun and Yang, Insoon},
	month = oct,
	year = {2022},
	note = {arXiv:2109.07120 [cs]},
	pages = {10065--10072},
}

@InProceedings{pmlr-v28-levine13,
  title     = {Guided Policy Search},
  author    = {Levine, Sergey and Koltun, Vladlen},
  booktitle = {Proceedings of the 30th International Conference on Machine Learning},
  pages     = {1--9},
  year      = {2013},
  editor    = {Dasgupta, Sanjoy and McAllester, David},
  volume    = {28},
  series    = {Proceedings of Machine Learning Research},
  address   = {Atlanta, Georgia, USA},
  month     = {17--19 Jun},
  publisher = {PMLR},
  url       = {https://proceedings.mlr.press/v28/levine13.html}
}

@inproceedings{kurtz_generative_2025,
	title = {Generative {Predictive} {Control}: {Flow} {Matching} {Policies} for {Dynamic}, {Difficult}-to-{Demonstrate} {Tasks}},
	author = {Kurtz, Vince and Burdick, Joel W.},
	booktitle = {2026 IEEE International Conference on Robotics and Automation (ICRA)},
	year = {2026},
}

@misc{kucuktabak_real-world_2026,
	title = {Real-{World} {Reinforcement} {Learning} with {MPC} {Scaffolding} for {Dexterous} {Manipulation}},
	url = {http://arxiv.org/abs/2609.14878},
	doi = {10.48550/arXiv.2609.14878},
	language = {en},
	urldate = {2026-09-21},
	publisher = {arXiv},
	author = {Küçüktabak, Emek Barış and Patel, Karankumar and Yang, Zhaodong and Cui, Jinda and Sasabuchi, Kazuhiro and Takamatsu, Jun},
	month = sep,
	year = {2026},
	note = {arXiv:2609.14878 [cs.RO]},
}

@article{reiter_synthesis_2026,
	title = {Synthesis of model predictive control and reinforcement learning: {Survey} and classification},
	volume = {61},
	issn = {13675788},
	shorttitle = {Synthesis of model predictive control and reinforcement learning},
	url = {https://linkinghub.elsevier.com/retrieve/pii/S1367578826000015},
	doi = {10.1016/j.arcontrol.2026.101045},
	language = {en},
	urldate = {2026-04-02},
	journal = {Annual Reviews in Control},
	author = {Reiter, Rudolf and Hoffmann, Jasper and Reinhardt, Dirk and Messerer, Florian and Baumgärtner, Katrin and Sawant, Shambhuraj and Bödecker, Joschka and Diehl, Moritz and Gros, Sebastien},
	year = {2026},
	pages = {101045},
}

@article{bogdanovic_model-free_2022,
	title = {Model-free reinforcement learning for robust locomotion using demonstrations from trajectory optimization},
	volume = {9},
	issn = {2296-9144},
	url = {https://www.frontiersin.org/articles/10.3389/frobt.2022.854212/full},
	doi = {10.3389/frobt.2022.854212},
	urldate = {2026-04-02},
	journal = {Frontiers in Robotics and AI},
	author = {Bogdanovic, Miroslav and Khadiv, Majid and Righetti, Ludovic},
	month = aug,
	year = {2022},
	pages = {854212},
}

@inproceedings{dawood_handling_2023,
	address = {London, United Kingdom},
	title = {Handling {Sparse} {Rewards} in {Reinforcement} {Learning} {Using} {Model} {Predictive} {Control}},
	copyright = {https://doi.org/10.15223/policy-029},
	isbn = {979-8-3503-2365-8},
	url = {https://ieeexplore.ieee.org/document/10161492/},
	doi = {10.1109/ICRA48891.2023.10161492},
	urldate = {2026-04-02},
	booktitle = {2023 {IEEE} {International} {Conference} on {Robotics} and {Automation} ({ICRA})},
	publisher = {IEEE},
	author = {Dawood, Murad and Dengler, Nils and De Heuvel, Jorge and Bennewitz, Maren},
	month = may,
	year = {2023},
	pages = {879--885},
}

@inproceedings{george_minimizing_2023,
	title = {Minimizing {Human} {Assistance}: {Augmenting} a {Single} {Demonstration} for {Deep} {Reinforcement} {Learning}},
	shorttitle = {Minimizing {Human} {Assistance}},
	url = {https://ieeexplore.ieee.org/document/10161119/},
	doi = {10.1109/ICRA48891.2023.10161119},
	urldate = {2026-05-05},
	booktitle = {2023 {IEEE} {International} {Conference} on {Robotics} and {Automation} ({ICRA})},
	author = {George, Abraham and Bartsch, Alison and Farimani, Amir Barati},
	month = may,
	year = {2023},
	pages = {5027--5033},
}

@inproceedings{nair_overcoming_2018,
	title = {Overcoming {Exploration} in {Reinforcement} {Learning} with {Demonstrations}},
	url = {https://ieeexplore.ieee.org/document/8463162/},
	doi = {10.1109/ICRA.2018.8463162},
	urldate = {2026-05-08},
	booktitle = {2018 {IEEE} {International} {Conference} on {Robotics} and {Automation} ({ICRA})},
	author = {Nair, Ashvin and McGrew, Bob and Andrychowicz, Marcin and Zaremba, Wojciech and Abbeel, Pieter},
	month = may,
	year = {2018},
	note = {ISSN: 2577-087X},
	pages = {6292--6299},
}

@article{jeon_residual_2025,
	title = {Residual {MPC}: {Blending} {Reinforcement} {Learning} with {GPU}-{Parallelized} {Model} {Predictive} {Control}},
	author = {Jeon, Se Hwan and Lee, Ho Jae and Hong, Seungwoo and Kim, Sangbae},
	journal = {IEEE Transactions on Robotics},
	volume = {42},
	pages = {3469--3488},
	year = {2026},
	doi = {10.1109/TRO.2026.3721669},
}

@article{cheng_rambo_2025,
	title = {Rambo: {RL}-{Augmented} {Model}-{Based} {Whole}-{Body} {Control} for {Loco}-{Manipulation}},
	volume = {10},
	copyright = {https://ieeexplore.ieee.org/Xplorehelp/downloads/license-information/IEEE.html},
	issn = {2377-3766, 2377-3774},
	shorttitle = {Rambo},
	url = {https://ieeexplore.ieee.org/document/11106746/},
	doi = {10.1109/LRA.2025.3594984},
	number = {9},
	urldate = {2026-04-02},
	journal = {IEEE Robotics and Automation Letters},
	author = {Cheng, Jin and Kang, Dongho and Fadini, Gabriele and Shi, Guanya and Coros, Stelian},
	month = sep,
	year = {2025},
	pages = {9462--9469},
}

@misc{unitree_rl_mjlab,
  author       = {{Unitree Robotics}},
  title        = {{Unitree RL mjlab}},
  year         = {2026},
  howpublished = {\url{https://github.com/unitreerobotics/unitree_rl_mjlab}},
  note         = {GitHub repository. Accessed: 2026-05-15}
}

@phdthesis{laud_theory_2004,
	address = {USA},
	type = {{PhD} {Thesis}},
	title = {Theory and application of reward shaping in reinforcement learning},
	school = {University of Illinois at Urbana-Champaign},
	author = {Laud, Adam Daniel},
	year = {2004},
}

@inproceedings{jeon_benchmarking_2023,
	title = {Benchmarking {Potential} {Based} {Rewards} for {Learning} {Humanoid} {Locomotion}},
	url = {http://arxiv.org/abs/2307.10142},
	doi = {10.1109/ICRA48891.2023.10160885},
	language = {en},
	urldate = {2025-12-10},
	booktitle = {2023 {IEEE} {International} {Conference} on {Robotics} and {Automation} ({ICRA})},
	author = {Jeon, Se Hwan and Heim, Steve and Khazoom, Charles and Kim, Sangbae},
	month = may,
	year = {2023},
	note = {arXiv:2307.10142 [cs]},
	pages = {9204--9210},
}

@inproceedings{malysheva_learning_2018,
	title = {Learning to {Run} with {Potential}-{Based} {Reward} {Shaping} and {Demonstrations} from {Video} {Data}},
	url = {http://arxiv.org/abs/2012.08824},
	doi = {10.1109/ICARCV.2018.8581310},
	language = {en},
	urldate = {2025-12-10},
	booktitle = {2018 15th {International} {Conference} on {Control}, {Automation}, {Robotics} and {Vision} ({ICARCV})},
	author = {Malysheva, Aleksandra and Kudenko, Daniel and Shpilman, Aleksei},
	month = nov,
	year = {2018},
	note = {arXiv:2012.08824 [cs]},
	pages = {286--291},
}

@article{harutyunyan_expressing_2015,
	title = {Expressing {Arbitrary} {Reward} {Functions} as {Potential}-{Based} {Advice}},
	volume = {29},
	issn = {2374-3468, 2159-5399},
	url = {https://ojs.aaai.org/index.php/AAAI/article/view/9628},
	doi = {10.1609/aaai.v29i1.9628},
	language = {en},
	number = {1},
	urldate = {2025-12-10},
	journal = {Proceedings of the AAAI Conference on Artificial Intelligence},
	author = {Harutyunyan, Anna and Devlin, Sam and Vrancx, Peter and Nowe, Ann},
	month = feb,
	year = {2015},
}

@inproceedings{westenbroek_lyapunov_2022,
	title = {Lyapunov {Design} for {Robust} and {Efficient} {Robotic} {Reinforcement} {Learning}},
	url = {https://proceedings.mlr.press/v205/westenbroek23a.html},
	language = {en},
	booktitle = {Proceedings of the 6th Conference on Robot Learning},
	author = {Westenbroek, Tyler and Castaneda, Fernando and Agrawal, Ayush and Sastry, Shankar and Sreenath, Koushil},
	year = {2023},
	series = {Proceedings of Machine Learning Research},
	volume = {205},
	pages = {2125--2135},
	publisher = {PMLR},
}

@inproceedings{yang_cbf-rl_2025,
	title = {{CBF}-{RL}: {Safety} {Filtering} {Reinforcement} {Learning} in {Training} with {Control} {Barrier} {Functions}},
	author = {Yang, Lizhi and Werner, Blake and de Sa, Massimiliano and Ames, Aaron D.},
	booktitle = {2026 IEEE International Conference on Robotics and Automation (ICRA)},
	year = {2026},
}

@article{ji_concurrent_2022,
	title = {Concurrent {Training} of a {Control} {Policy} and a {State} {Estimator} for {Dynamic} and {Robust} {Legged} {Locomotion}},
	volume = {7},
	issn = {2377-3766, 2377-3774},
	url = {http://arxiv.org/abs/2202.05481},
	doi = {10.1109/LRA.2022.3151396},
	language = {en},
	number = {2},
	urldate = {2025-12-10},
	journal = {IEEE Robotics and Automation Letters},
	author = {Ji, Gwanghyeon and Mun, Juhyeok and Kim, Hyeongjun and Hwangbo, Jemin},
	month = apr,
	year = {2022},
	note = {arXiv:2202.05481 [cs]},
	pages = {4630--4637},
}

@article{lee_learning_2020,
	title = {Learning {Quadrupedal} {Locomotion} over {Challenging} {Terrain}},
	volume = {5},
	issn = {2470-9476},
	url = {http://arxiv.org/abs/2010.11251},
	doi = {10.1126/scirobotics.abc5986},
	language = {en},
	number = {47},
	urldate = {2025-12-10},
	journal = {Science Robotics},
	author = {Lee, Joonho and Hwangbo, Jemin and Wellhausen, Lorenz and Koltun, Vladlen and Hutter, Marco},
	month = oct,
	year = {2020},
	note = {arXiv:2010.11251 [cs]},
	pages = {eabc5986},
}

@inproceedings{ng_policy_1999,
	address = {San Francisco, CA, USA},
	series = {{ICML} '99},
	title = {Policy {Invariance} {Under} {Reward} {Transformations}: {Theory} and {Application} to {Reward} {Shaping}},
	isbn = {1-55860-612-2},
	booktitle = {Proceedings of the {Sixteenth} {International} {Conference} on {Machine} {Learning}},
	publisher = {Morgan Kaufmann Publishers Inc.},
	author = {Ng, Andrew Y. and Harada, Daishi and Russell, Stuart J.},
	year = {1999},
	pages = {278--287},
}

@article{liang_learning_2026,
	title = {Learning robust bipedal running via structured gait and trajectory guidance},
	copyright = {https://www.cambridge.org/core/terms},
	issn = {0263-5747, 1469-8668},
	url = {https://www.cambridge.org/core/product/identifier/S0263574725103007/type/journal_article},
	doi = {10.1017/S0263574725103007},
	language = {en},
	urldate = {2026-01-19},
	journal = {Robotica},
	author = {Liang, Yunpeng and Peng, Zhihui and Zhao, Yanzheng and Yan, Weixin},
	month = jan,
	year = {2026},
	pages = {150--168},
	volume = {44},
	number = {1},
}

@article{muller_olaf_2025,
	title = {Olaf: {Bringing} an {Animated} {Character} to {Life} in the {Physical} {World}},
	shorttitle = {Olaf},
	url = {https://doi.org/10.1109/LRA.2026.3685938},
	doi = {10.1109/LRA.2026.3685938},
	language = {en},
	author = {Müller, David and Knoop, Espen and Mylonopoulos, Dario and Serifi, Agon and Hopkins, Michael A. and Grandia, Ruben and Bächer, Moritz},
	month = jun,
	year = {2026},
	journal = {IEEE Robotics and Automation Letters},
	volume = {11},
	number = {6},
	pages = {7580--7587},
}

@inproceedings{xue_opening_2025,
	title = {Opening the {Sim}-to-{Real} {Door} for {Humanoid} {Pixel}-to-{Action} {Policy} {Transfer}},
	url = {https://openaccess.thecvf.com/content/CVPR2026/html/Xue_Opening_the_Sim-to-Real_Door_for_Humanoid_Pixel-to-Action_Policy_Transfer_CVPR_2026_paper.html},
	language = {en},
	author = {Xue, Haoru and He, Tairan and Wang, Zi and Ben, Qingwei and Xiao, Wenli and Luo, Zhengyi and Da, Xingye and Castañeda, Fernando and Shi, Guanya and Sastry, Shankar and Fan, Linxi and Zhu, Yuke},
	month = jun,
	year = {2026},
	booktitle = {Proceedings of the IEEE/CVF Conference on Computer Vision and Pattern Recognition (CVPR)},
	pages = {6642--6652},
}

@inproceedings{rudin_learning_2022,
	title = {Learning to {Walk} in {Minutes} {Using} {Massively} {Parallel} {Deep} {Reinforcement} {Learning}},
	url = {https://proceedings.mlr.press/v164/rudin22a.html},
	language = {en},
	urldate = {2026-04-02},
	booktitle = {Proceedings of the 5th {Conference} on {Robot} {Learning}},
	publisher = {PMLR},
	author = {Rudin, Nikita and Hoeller, David and Reist, Philipp and Hutter, Marco},
	month = jan,
	year = {2022},
	note = {ISSN: 2640-3498},
	pages = {91--100},
}

@inproceedings{siekmann_sim--real_2021,
	address = {Xi'an, China},
	title = {Sim-to-{Real} {Learning} of {All} {Common} {Bipedal} {Gaits} via {Periodic} {Reward} {Composition}},
	copyright = {https://ieeexplore.ieee.org/Xplorehelp/downloads/license-information/IEEE.html},
	isbn = {978-1-7281-9077-8},
	url = {https://ieeexplore.ieee.org/document/9561814/},
	doi = {10.1109/ICRA48506.2021.9561814},
	urldate = {2026-04-02},
	booktitle = {2021 {IEEE} {International} {Conference} on {Robotics} and {Automation} ({ICRA})},
	publisher = {IEEE},
	author = {Siekmann, Jonah and Godse, Yesh and Fern, Alan and Hurst, Jonathan},
	month = may,
	year = {2021},
	pages = {7309--7315},
}

@article{kim_not_2024,
	title = {Not {Only} {Rewards} but {Also} {Constraints}: {Applications} on {Legged} {Robot} {Locomotion}},
	volume = {40},
	copyright = {https://ieeexplore.ieee.org/Xplorehelp/downloads/license-information/IEEE.html},
	issn = {1552-3098, 1941-0468},
	shorttitle = {Not {Only} {Rewards} but {Also} {Constraints}},
	url = {https://ieeexplore.ieee.org/document/10530429/},
	doi = {10.1109/TRO.2024.3400935},
	urldate = {2026-04-02},
	journal = {IEEE Transactions on Robotics},
	author = {Kim, Yunho and Oh, Hyunsik and Lee, Jeonghyun and Choi, Jinhyeok and Ji, Gwanghyeon and Jung, Moonkyu and Youm, Donghoon and Hwangbo, Jemin},
	year = {2024},
	pages = {2984--3003},
}

@misc{unitree_go2,
  author       = {{Unitree Robotics}},
  title        = {{Unitree Go2}},
  howpublished = {\url{https://www.unitree.com/go2/}},
  year         = {n.d.},
  note         = {Product page. Accessed: 2026-05-17}
}

@misc{howell2022,
  title={{Predictive Sampling: Real-time Behaviour Synthesis with MuJoCo}},
  author={Howell, Taylor and Gileadi, Nimrod and Tunyasuvunakool, Saran and Zakka, Kevin and Erez, Tom and Tassa, Yuval},
  archivePrefix={arXiv},
  eprint={2212.00541},
  primaryClass={cs.RO},
  url={https://arxiv.org/abs/2212.00541},
  doi={10.48550/arXiv.2212.00541},
  year={2022},
  month={Dec}
}

@inproceedings{schramm_reference-free_2025,
	title = {Reference-{Free} {Sampling}-{Based} {Model} {Predictive} {Control}},
	url = {https://hal.science/hal-05401976},
	language = {en},
	author = {Schramm, Fabian and Fabre, Pierre and Perrin-Gilbert, Nicolas and Carpentier, Justin},
	year = {2026},
	booktitle = {IEEE International Conference on Robotics and Automation (ICRA)},
}

@article{amatucci_primal-dual_2026,
	title = {Primal-{Dual} {iLQR} for {GPU}-{Accelerated} {Learning} and {Control} in {Legged} {Robots}},
	volume = {11},
	copyright = {https://ieeexplore.ieee.org/Xplorehelp/downloads/license-information/IEEE.html},
	issn = {2377-3766, 2377-3774},
	url = {https://ieeexplore.ieee.org/document/11248841/},
	doi = {10.1109/LRA.2025.3632610},
	number = {1},
	urldate = {2026-04-02},
	journal = {IEEE Robotics and Automation Letters},
	author = {Amatucci, Lorenzo and Sousa-Pinto, João and Turrisi, Giulio and Orban, Dominique and Barasuol, Victor and Semini, Claudio},
	month = jan,
	year = {2026},
	pages = {1010--1017},
}

\appendix
\section{Problem Statement}
\label{appendix:prob_statement}
This section introduces the two main theoretical components of MPC-Injection, the MPC controller and the off-policy RL algorithm. 
We formulate MPC and RL following~\cite{reiter_synthesis_2026}. In our experiments they act in the same controlled transition system but optimize different objectives; we use the following MDP notation to explain how controller transitions enter off-policy RL. An MDP is defined by a 5-tuple
\begin{equation}
    \mathcal{M} := (\mathcal{S}, \mathcal{A}, P, l, \gamma)
\end{equation}
with state space $\mathcal{S}$, action space $\mathcal{A}$, state transition model $P:\mathcal{S} \times \mathcal{A} \rightarrow \text{Dist}(\mathcal{S})$, stage cost or reward $l: \mathcal{S} \times \mathcal{A} \rightarrow \mathbb{R}$, and discount factor $\gamma \in (0, 1]$. Given a policy $\pi$, we define the value function 
\begin{equation}
    V^\pi (s) := \mathbb{E} \left[ \sum_{k=0}^\infty \gamma^k l(S_k, A_k ) \;|\; S_0 = s \right]
    \label{eq:state-val_fxn_rl}
\end{equation}
and the action-value function
\begin{equation}
    Q^\pi (s,a) := \mathbb{E} \left[ \sum_{k=0}^\infty \gamma^k l(S_k, A_k ) \;|\; S_0 = s, A_0 = a\right]
    \label{eq:action-state_val_fxn_rl}
\end{equation}
Solving an MDP means finding an optimal policy $\pi^*$ that optimizes the total expected return $J(\pi)=\mathbb{E}_{\tau\sim\pi}[R(\tau)]$, where $\tau = (S_0, A_0, S_1, A_1, \dots, S_n)$ is a trajectory and $R(\tau)=\sum_{k=0}^{n-1} \gamma^k l(S_k, A_k)$ is its total cost or reward. RL maximizes the task return, whereas MPC minimizes its separate finite-horizon controller cost.

\subsection{Model Predictive Control (MPC)}
Through the MDP formulation, the MPC policy is
\begin{equation}
    \pi^\text{MPC} (s) = \arg \min_{a\in A} \;Q^{\text{MPC}}(s,a)
\end{equation}
where the action-value function for the MPC is defined as
\begin{equation}
\begin{aligned}
Q^{\mathrm{MPC}}(s,a)
  &= \min_{x_{1:N},\, u_{1:N-1}} \sum_{k=0}^{N-1} l^{\mathrm{MPC}}(x_k,u_k)
     + \bar{V}^{\mathrm{MPC}}(x_N) \\
\text{s.t.}\quad
  &x_0 = s, \qquad u_0 = a, \\
  &x_{k+1} = f^{\mathrm{MPC}}(x_k,u_k), \quad 0 \le k < N.
\end{aligned}
\label{eq:action-state_val_fxn_mpc}
\end{equation}
Here, $N \in \mathbb{N}$ is the MPC prediction horizon, $x_k$ and $u_k$ are the predicted state and control at stage $k$, $l^{\mathrm{MPC}}$ is the stage cost, $\bar{V}^{\mathrm{MPC}}$ is the terminal cost or terminal value approximation, and $f^{\mathrm{MPC}}$ is the deterministic prediction model that approximates the environment transition distribution $P(\cdot \mid s,a)$. At inference time, MPC repeatedly solves this finite-horizon problem at the current state, applies only the first control input, observes the next state, and repeats in a receding-horizon manner. Under this cost-minimization convention, lower values of $Q^{\mathrm{MPC}}(s,a)$ are preferred. The MPC controller can be of any type, gradient-based~\cite{kim_contact-implicit_2025, du_gato_2025, alavilli_tinympc_2025} or sampling-based~\cite{schramm_reference-free_2025, howell2022}, as long as the trajectories collected from it are from the same RL environment that the off-policy algorithm learns from. We demonstrate this in Section~\ref{sec:experiments}.

\subsection{Off-Policy RL}
In deep RL the policy $\pi^{\text{RL}}_\theta$ is a neural network with weights $\theta$. The optimal policy in off-policy actor-critic methods, such as Soft Actor-Critic (SAC)~\cite{haarnoja_soft_2018} or Twin Delayed Deep Deterministic Policy Gradient (TD3)~\cite{fujimoto_addressing_2018}, is found by solving
\begin{equation}
    \max_\theta \mathbb{E}_{s \sim \mathcal{D}} \left[ Q_\phi(s, \pi^{\text{RL}}_\theta(s)) \right]
\end{equation}

where $\mathcal{D}$ is a replay buffer and $\phi$ are the weights of the learned Q-function. Off-policy methods can learn from transitions generated by any policy stored in the replay buffer. Each of the off-policy algorithms we use follows the general steps in Algorithm~\ref{alg:off_policy_mpc_injection}.

\begin{algorithm}[ht]
\caption{Off-Policy RL with MPC-Injection.\\\textcolor{blue}{The blue lines are the only modification needed to recover the MPC-Injection variant.}}
\label{alg:off_policy_mpc_injection}
\begin{algorithmic}[1]
\State \textbf{Input:} initial policy parameters $\theta$, Q-function parameters $\phi$, empty replay buffer $\mathcal{D}$, \textcolor{blue}{MPC replay buffer percentage $p$}, \textcolor{blue}{MPC transitions}
\State Set target parameters equal to main parameters
\Repeat
    \State Observe state $s$ and select action $a$
    \State Execute $a$ in the environment
    \State Observe next state $s'$, reward $r$, and done signal $d$
    \If{$0\% \leq p < 100\%$}
        \State Store $(s,a,r,s',d)$ in replay buffer $\mathcal{D}$
        \State \textcolor{blue}{Store MPC transitions in $\mathcal{D}$ until ratio $p$ is satisfied}
    \Else \Comment{$p=100\%$}
        \State \textcolor{blue}{Discard the online transition and store only the next MPC transition}
    \EndIf
    \If{$s'$ is terminal}
        \State Reset environment state
    \EndIf
    \If{it is time to update}
        \For{each update step}
            \State Randomly sample a batch of transitions, $\{(s,a,r,s',d)\}$, from $\mathcal{D}$
            \State Compute targets
            \State Update Q-function by one step of gradient descent
            \State Update policy by one step of gradient ascent
            \State Update target networks
        \EndFor
    \EndIf
\Until{convergence}
\end{algorithmic}
\end{algorithm}

Because off-policy methods can learn from transitions collected by different behavior policies, we mix trajectories from an MPC controller with trajectories from the learned policy in the same buffer. The rollouts are executed in the RL simulator, so internal MPC model mismatch can affect trajectory quality and behavior without by itself invalidating the resulting simulator transitions; the quadruped action conversion remains approximate as detailed in Appendix~\ref{appendix:converting_quadruped_mpc_to_rl}. MPC-Injection leverages this to perform behavior biasing. Additional details that differentiate the off-policy algorithms, particularly in the update steps, are described in the TD3 and SAC papers.

\FloatBarrier

\section{MPC Details} \label{appendix:mpc_details}

\textbf{Walker trajectory generation.} 
Walker trajectories were generated using the sampling-based model predictive controller from~\cite{howell2022}. The planner operated with a horizon of $0.8\,\mathrm{s}$ and a control update period of $0.025\,\mathrm{s}$, corresponding to
\[
N_{\mathrm{plan}} = \frac{0.8}{0.025} + 1 = 33
\]
planning nodes.
The MuJoCo simulator timestep was $0.0025\,\mathrm{s}$, so each MPC action was held for 10 simulator steps between planner updates. Each trajectory rollout had length 1000 simulator steps, corresponding to $2.5\,\mathrm{s}$ of simulated time, with one MPC optimization iteration per planner update.

The walker cost function used weights $w_{\mathrm{speed}}=1.0$, $w_{\mathrm{height}}=10.0$, $w_{\mathrm{rotation}}=3.0$, and $w_{\mathrm{control}}=0.1$, with target forward speed $v_x^\star = 1.0\,\mathrm{m/s}$ and target torso height $h^\star = 1.2\,\mathrm{m}$. Initial states were generated to be similar to the RL environment by scanning a grid over the admissible walker joint ranges with grid spacing $0.1$, with the pitch root angle sampled over $[-\pi,\pi]$, root translation coordinates fixed at zero, and all initial joint velocities zero.

\textbf{Quadruped trajectory generation.}
Quadruped trajectories were generated using the MPX model predictive controller~\cite{amatucci_primal-dual_2026} in its Go2 whole-body iLQR configuration. The controller operated at $50\,\mathrm{Hz}$ with planning timestep $0.02\,\mathrm{s}$ and horizon length $N=25$, corresponding to a prediction horizon of $0.5\,\mathrm{s}$. Trajectory rollouts were executed in MuJoCo at $200\,\mathrm{Hz}$, so the MPC action was updated every 4 simulator steps. Each rollout had length 1000 control steps, corresponding to $20\,\mathrm{s}$ of simulated time. The solver was warm-started between successive MPC calls by shifting the previous state, control, and dual trajectories forward.

The gait generator used a trotting pattern with duty factor $0.5$, step frequency $2.0\,\mathrm{Hz}$, step height $0.065\,\mathrm{m}$, and leg phase offsets $[0.5,\,0.0,\,0.0,\,0.5]$. When the commanded speed magnitude fell below a small threshold, the controller switched to standing mode by setting the duty factor to $1.0$.

The stage cost penalized base position, base orientation, joint angles, base linear and angular velocity, joint velocity, swing-foot position, joint torque, and ground reaction force, with diagonal weights
\[
Q_p = \mathrm{diag}(0,0,10^4),\quad
Q_{\mathrm{rot}} = \mathrm{diag}(10^3,10^4,10^3),\quad
Q_q = 10^{-1}I,\quad
Q_{\dot p} = 2\times 10^3 I,
\]
\[
Q_{\omega} = 5\times 10^2 I,\quad
Q_{\dot q} = 10^{-1}I,\quad
Q_{\tau} = 10^{-1}I,\quad
Q_{\mathrm{grf}} = 10^{-2}I,
\]
together with a swing-foot tracking weight
\[
Q_{\mathrm{foot}} = \mathrm{diag}\!\bigl([10^4,\,10^4,\,10^5]^{\otimes 4}\bigr).
\]
The optimizer used a Gauss--Newton Hessian approximation and included penalties for friction-cone, torque-limit, and joint-speed violations. The applied simulator torque combined the MPC output with a low-level PD controller ($K_p=10$, $K_d=2$):
\[
\tau_{\mathrm{app}}
=
\tau_{\mathrm{mpc}}
+
10\,(q^{\mathrm{des}}-q)
-
2\,\dot q.
\]
Initial states and commands were sampled to match the RL training environment. Joint positions were perturbed within $\pm 0.05\,\mathrm{rad}$, base roll and pitch within $\pm 0.03\,\mathrm{rad}$, and joint velocities within $\pm 0.05\,\mathrm{rad/s}$. Forward velocity commands were sampled from $[0,\,0.5]\,\mathrm{m/s}$, while lateral and yaw-rate commands were fixed at zero. Commands were resampled every 250 control steps (5 seconds at 50\,Hz). Trajectories were discarded if the robot fell, defined as roll or pitch exceeding $0.5\,\mathrm{rad}$ or base height dropping below $0.1\,\mathrm{m}$.

\section{Details on Policy Learning with Off-Policy Reinforcement Learning} \label{appendix:training_details}
We use the TD3 and SAC implementations from Stable-Baselines3~\cite{stable-baselines3} as the baseline off-policy RL algorithms, with MPC-Injection implemented in the same framework. Table~\ref{tab:training_hyperparams_walker_quadruped} lists the training hyperparameters.

The MLP actor and critic each use two hidden layers of size 256 for both the walker and quadruped tasks, with input and output dimensions matched to each task. The walker actor and critic receive the same environment observation and the actor outputs the continuous action vector. For the quadruped we use an asymmetric actor-critic architecture: the actor receives a 45-dimensional policy observation containing hardware-available signals and outputs a 12-dimensional vector of joint position residuals, while the critic receives the same 45-dimensional observation concatenated with a 3-dimensional privileged observation of base linear velocity (48 dimensions total).

All hidden layers use ReLU activations. The actor and critic networks are optimized with Adam using the constant learning rates in Table~\ref{tab:training_hyperparams_walker_quadruped}; the remaining Adam settings are the PyTorch defaults ($\beta_1=0.9$, $\beta_2=0.999$, $\epsilon=10^{-8}$, and zero weight decay). We use the Stable-Baselines3 default step-based training frequency of one vectorized environment step. After each collection step, the number of optimizer steps is set by the ``Grad'' column of Table~\ref{tab:training_hyperparams_walker_quadruped}: a value of 1 performs one optimizer step, while $-1$ performs as many optimizer steps as environment transitions collected in that rollout (here, $N_{\mathrm{env}}$ transitions).

\subsection{Quadruped Reward Functions} \label{appendix:reward_fxns}

This subsection details the reward and penalty terms for the three quadruped reward configurations used in the main paper. The MPC-Injection simple reward is in Table~\ref{tab:simple_reward_active}. The AMP-MPC baseline is in Tables~\ref{tab:go2_amp_reward_composition}, \ref{tab:go2_amp_reward_positive}, and \ref{tab:go2_amp_reward_penalty}. The reward-shaping baseline is in Tables~\ref{tab:full_reward_positive} and \ref{tab:full_reward_penalty}. We found in testing that the forward linear velocity term was necessary for off-policy methods to learn to walk under the simple reward or the reward-shaping baseline. On-policy methods, e.g., PPO, did not require this term.

\begin{table*}[!t]
\centering
\scriptsize
\renewcommand{\arraystretch}{1.12}
\setlength{\tabcolsep}{4pt}
\begin{tabularx}{\textwidth}{@{}llcccccccX@{}}
\toprule
\textbf{Env.} & \textbf{Algo} & \textbf{Steps} & \textbf{$N_{\text{env}}$} & \textbf{LR} & \textbf{Buffer} & \textbf{Start} & \textbf{Batch} & \textbf{Grad} & \textbf{Other training settings} \\
\midrule

Walker & SAC
& $5\times 10^5$ & 4 & $3\times 10^{-4}$ & $10^6$ & $10^4$ & 256 & 1
& $\tau=0.005$, $\gamma=0.99$ \\

Walker & TD3
& $5\times 10^5$ & 4 & $3\times 10^{-4}$ & $10^6$ & $10^4$ & 256 & $-1$
& $\tau=0.005$, $\gamma=0.99$, $\pi$ delay $=2$ \\

Walker & SAC-MPC
& $5\times 10^5$ & 4 & $3\times 10^{-4}$ & $10^6$ & $10^4$ & 256 & 1
& $\tau=0.005$, $\gamma=0.99$ \\

Walker & TD3-MPC
& $5\times 10^5$ & 4 & $3\times 10^{-4}$ & $10^6$ & $10^4$ & 256 & $-1$
& $\tau=0.005$, $\gamma=0.99$, $\pi$ delay $=2$ \\

Quadruped & SAC
& $10^6$ & 256 & $3\times 10^{-4}$ & $5\times 10^6$ & $5\times 10^4$ & 256 & $-1$
& $\tau=0.005$, $\gamma=0.99$ \\


Quadruped & SAC-MPC
& $10^6$ & 256 & $3\times 10^{-4}$ & $5\times 10^6$ & $5\times 10^4$ & 256 & $-1$
& $\tau=0.005$, $\gamma=0.99$ \\

\bottomrule
\end{tabularx}
\vspace{-4pt}
\caption{Training hyperparameters for the walker and quadruped tasks and the RL algorithms used.}
\label{tab:training_hyperparams_walker_quadruped}
\end{table*}

\begin{table}[!t]
\centering
\scriptsize
\renewcommand{\arraystretch}{1.15}
\setlength{\tabcolsep}{4pt}

\begin{tabularx}{\linewidth}{@{}%
    >{\raggedright\arraybackslash}p{0.22\linewidth}
    >{\centering\arraybackslash}X
    >{\centering\arraybackslash}p{0.14\linewidth}
@{}}
\toprule
\textbf{MPC-Injection Quadruped Reward Terms} & \textbf{Definition} & \textbf{Scale} \\
\midrule

Linear velocity-tracking reward &
$\displaystyle
r_\text{track}(\mathbf{v}^\text{cmd}) =
\exp\!\left(
-\frac{
\|\mathbf{v}_{xy}^\text{cmd}-\mathbf{v}_{xy}^{\text{base}}\|_2^2
+ 2(v_z^\text{base})^2
}{0.25}
\right)
$ &
$\displaystyle w_\text{track}=1.5$ \\

Forward linear velocity reward &
$\displaystyle
r_\text{fwd}(\mathbf{v}^\text{cmd})=
\begin{cases}
\operatorname{clip}\!\left(
(\mathbf{v}_{xy}^{\text{base}})^\top
\frac{\mathbf{v}_{xy}^\text{cmd}}{\|\mathbf{v}_{xy}^\text{cmd}\|_2},
\,0,\,
\|\mathbf{v}_{xy}^\text{cmd}\|_2
\right),
& \|\mathbf{v}_{xy}^\text{cmd}\|_2 > 0.1, \\[4pt]
0, & \text{otherwise.}
\end{cases}
$ &
$\displaystyle w_\text{fwd}=1.5$ \\

\bottomrule
\end{tabularx}
\vspace{2pt}
\caption{Reward terms for the simple quadruped velocity tracking task used to showcase MPC-Injection's behavior shaping abilities.}
\label{tab:simple_reward_active}
\end{table}

\begin{table}[!t]
    \centering
    \scriptsize
    \renewcommand{\arraystretch}{1.15}
    \setlength{\tabcolsep}{4pt}

    \begin{tabularx}{\linewidth}{@{}%
        >{\raggedright\arraybackslash}p{0.26\linewidth}
        >{\centering\arraybackslash}X
        >{\centering\arraybackslash}p{0.16\linewidth}
    @{}}
    \toprule
    \textbf{AMP-MPC Reward Component} & \textbf{Definition} & \textbf{Scale} \\
    \midrule

    AMP style reward &
    $\displaystyle
    r_{\mathrm{style}} =
    \alpha_{\mathrm{AMP}}
    \max\!\left(
    1-\frac{1}{4}\bigl(D(s_t,s_{t+1})-1\bigr)^2,\,
    0
    \right)
    $ &
    $\displaystyle \alpha_{\mathrm{AMP}} = 2.0$ \\

    Task reward aggregation &
    $\displaystyle
    r_{\mathrm{task}} =
    \max\!\left(
    1.5\,r_{\mathrm{lin}}
    +0.5\,r_{\mathrm{yaw}}
    -3\times 10^{-4}\,c_{\mathrm{act}},
    \,0
    \right)
    $ &
    clipped at zero \\

    Final PPO reward &
    $\displaystyle
    r_{\mathrm{PPO}} =
    (1-\lambda_{\mathrm{task}})\,r_{\mathrm{style}}
    +\lambda_{\mathrm{task}}\,r_{\mathrm{task}}
    $ &
    $\displaystyle \lambda_{\mathrm{task}} = 0.3$ \\

    \bottomrule
    \end{tabularx}
    \vspace{2pt}
    \caption{Reward terms used for the AMP-MPC baseline. Here $D(s_t,s_{t+1})$ is the AMP discriminator output and $r_{\mathrm{task}}$ is the environment task reward.}
    \label{tab:go2_amp_reward_composition}
\end{table}

\begin{table}[!t]
    \centering
    \scriptsize
    \renewcommand{\arraystretch}{1.15}
    \setlength{\tabcolsep}{4pt}

    \begin{tabularx}{\linewidth}{@{}%
        >{\raggedright\arraybackslash}p{0.27\linewidth}
        >{\centering\arraybackslash}X
        >{\centering\arraybackslash}p{0.15\linewidth}
    @{}}
    \toprule
    \textbf{AMP-MPC Positive Task-Reward Term} & \textbf{Definition} & \textbf{Scale} \\
    \midrule

    Linear velocity tracking &
    $\displaystyle
    r_{\mathrm{lin}} =
    \exp\!\left(
    -\frac{
    \|\mathbf{v}^{\mathrm{cmd}}_{xy}
    -\mathbf{v}^{\mathrm{base}}_{xy}\|_2^2
    }{0.25}
    \right)
    $ &
    $\displaystyle w_{\mathrm{lin}} = 1.5$ \\

    Yaw-rate tracking &
    $\displaystyle
    r_{\mathrm{yaw}} =
    \exp\!\left(
    -\frac{
    (\omega_z^{\mathrm{cmd}}-\omega_z^{\mathrm{base}})^2
    }{0.25}
    \right)
    $ &
    $\displaystyle w_{\mathrm{yaw}} = 0.5$ \\

    \bottomrule
    \end{tabularx}
    \vspace{2pt}
    \caption{Positive task-reward terms for the AMP-MPC baseline.}
    \label{tab:go2_amp_reward_positive}
\end{table}

\begin{table}[!t]
    \centering
    \scriptsize
    \renewcommand{\arraystretch}{1.15}
    \setlength{\tabcolsep}{4pt}

    \begin{tabularx}{\linewidth}{@{}%
        >{\raggedright\arraybackslash}p{0.27\linewidth}
        >{\centering\arraybackslash}X
        >{\centering\arraybackslash}p{0.15\linewidth}
    @{}}
    \toprule
    \textbf{AMP-MPC Task Penalty Term} & \textbf{Definition} & \textbf{Scale} \\
    \midrule

    Action rate penalty &
    $\displaystyle
    c_{\mathrm{act}} =
    \|\mathbf{a}_t-\mathbf{a}_{t-1}\|_2^2
    $ &
    $\displaystyle w_{\mathrm{act}} = -3\times10^{-4}$ \\

    \bottomrule
    \end{tabularx}
    \vspace{2pt}
    \caption{Task penalty term for the AMP-MPC baseline.}
    \label{tab:go2_amp_reward_penalty}
\end{table}

\begin{table}[!t]
    \centering
    \scriptsize
    \renewcommand{\arraystretch}{1.15}
    \setlength{\tabcolsep}{4pt}

    \begin{tabularx}{\linewidth}{@{}%
        >{\raggedright\arraybackslash}p{0.24\linewidth}
        >{\centering\arraybackslash}X
        >{\centering\arraybackslash}p{0.13\linewidth}
    @{}}
    \toprule
    \textbf{Reward Shaping Quadruped Positive Reward Term} & \textbf{Definition} & \textbf{Scale} \\
    \midrule

    Linear velocity tracking &
    $\displaystyle
    r_{\text{track-lin}} =
    \exp\!\left(
    -\frac{
    \|\mathbf{v}^{\text{cmd}}_{xy}-\mathbf{v}^{\text{base}}_{xy}\|_2^2
    +2(v_z^{\text{base}})^2
    }{0.25}
    \right)
    $ &
    $\displaystyle w_{\text{track-lin}} = 4.0$ \\

    Angular velocity tracking &
    $\displaystyle
    r_{\text{track-ang}} =
    \exp\!\left(
    -\frac{
    (\omega_z^{\text{cmd}}-\omega_z^{\text{base}})^2
    +0.05\|\boldsymbol{\omega}^{\text{base}}_{xy}\|_2^2
    }{0.25}
    \right)
    $ &
    $\displaystyle w_{\text{track-ang}} = 2.5$ \\

    Forward linear velocity reward &
    $\displaystyle
    r_{\text{fwd-lin}}=
    \begin{cases}
    \operatorname{clip}\!\left(
    (\mathbf{v}^{\text{base}}_{xy})^\top
    \frac{\mathbf{v}^{\text{cmd}}_{xy}}{\|\mathbf{v}^{\text{cmd}}_{xy}\|_2},
    \,0,\,
    \|\mathbf{v}^{\text{cmd}}_{xy}\|_2
    \right),
    & \|\mathbf{v}^{\text{cmd}}_{xy}\|_2 > 0.1, \\[4pt]
    0, & \text{otherwise,}
    \end{cases}
    $ &
    $\displaystyle w_{\text{fwd-lin}} = 6.0$ \\

    Forward yaw-rate reward &
    $\displaystyle
    r_{\text{fwd-ang}}=
    \begin{cases}
    \operatorname{clip}\!\left(
    \omega_z^{\text{base}}\,\operatorname{sign}(\omega_z^{\text{cmd}}),
    \,0,\,
    |\omega_z^{\text{cmd}}|
    \right),
    & |\omega_z^{\text{cmd}}| > 0.1, \\[4pt]
    0, & \text{otherwise,}
    \end{cases}
    $ &
    $\displaystyle w_{\text{fwd-ang}} = 1.0$ \\

    Variable posture reward &
    $\displaystyle
    r_{\text{pose}}=
    \exp\!\left(
    -\frac{1}{12}\sum_{j=1}^{12}
    \frac{(q_j-q_{j,0})^2}{\sigma_j(c)^2}
    \right)
    $ &
    $\displaystyle w_{\text{pose}} = 0.42$ \\

    Base height tracking &
    $\displaystyle
    r_{\text{height}}=
    \exp\!\left(
    -\frac{(z^{\text{base}}-0.27)^2}{0.01}
    \right)
    $ &
    $\displaystyle w_{\text{height}} = 1.0$ \\

    Feet air-time reward &
    $\displaystyle
    r_{\text{air}}=
    \chi\,
    \max\!\left(T_{\text{air}}-\left|t_{\text{mode}}-T_{\text{air}}\right|,\,0\right),
    \qquad
    T_{\text{air}}=0.245
    $ &
    $\displaystyle w_{\text{air}} = 0.75$ \\

    Scheduled diagonal gait reward &
    $\displaystyle
    r_{\text{gait}}=
    \chi\,
    \max\!\left(
    2\left(
    \frac{1}{4}\sum_{i=1}^{4}\mathbbold{1}[I_i=S_i(\phi)]
    -\frac{1}{2}
    \right),\,0
    \right)
    $ &
    $\displaystyle w_{\text{gait}} = 1.35$ \\

    \bottomrule
    \end{tabularx}
    \vspace{2pt}
    \caption{Positive reward terms for the fully reward-shaped quadruped velocity-tracking task. Here $c=\|\mathbf{v}^{\text{cmd}}_{xy}\|_2+|\omega_z^{\text{cmd}}|$, $\chi=\mathbbold{1}[c>0.1]$, and $I_i=\mathbbold{1}[i\in\text{contact}]$. For the air-time term, $t_{\text{mode}}=\min_i(I_i t_i^{\text{contact}}+(1-I_i)t_i^{\text{air}})$ when exactly two feet are in contact, and $t_{\text{mode}}=0$ otherwise. For the gait term, $\phi=(t/0.52)\bmod 1$, $S_i(\phi)=\mathbbold{1}[((\phi+\delta_i)\bmod 1)<0.52]$, with $\delta_{\mathrm{FR}}=\delta_{\mathrm{RL}}=0$ and $\delta_{\mathrm{FL}}=\delta_{\mathrm{RR}}=0.5$.}
    \label{tab:full_reward_positive}
\end{table}

\begin{table}[!t]
    \centering
    \scriptsize
    \renewcommand{\arraystretch}{1.15}
    \setlength{\tabcolsep}{4pt}

    \begin{tabularx}{\linewidth}{@{}%
        >{\raggedright\arraybackslash}p{0.24\linewidth}
        >{\centering\arraybackslash}X
        >{\centering\arraybackslash}p{0.13\linewidth}
    @{}}
    \toprule
    \textbf{Reward Shaping Quadruped Penalty Term} & \textbf{Definition} & \textbf{Scale} \\
    \midrule

    Lateral velocity penalty &
    $\displaystyle
    c_{\text{lat}}=
    \chi\,(v_y^{\text{base}})^2
    $ &
    $\displaystyle w_{\text{lat}} = -1.0$ \\

    Flat orientation penalty &
    $\displaystyle
    c_{\text{flat}} = \|\mathbf{g}^{\text{proj}}_{xy}\|_2^2
    $ &
    $\displaystyle w_{\text{flat}} = -0.7$ \\

    Body angular velocity penalty &
    $\displaystyle
    c_{\text{body-ang}}=
    \|\boldsymbol{\omega}^{\text{world}}_{xy}\|_2^2
    $ &
    $\displaystyle w_{\text{body-ang}} = -0.16$ \\

    Pitch tilt penalty &
    $\displaystyle
    c_{\text{pitch}}=
    \theta_{\text{pitch}}^2
    $ &
    $\displaystyle w_{\text{pitch}} = -2.0$ \\

    Angular momentum penalty &
    $\displaystyle
    c_{\text{angmom}}=
    \|\mathbf{h}_{\text{base}}\|_2^2
    $ &
    $\displaystyle w_{\text{angmom}} = -0.014$ \\

    Termination penalty &
    $\displaystyle
    c_{\text{term}}= \mathbbold{1}[\text{terminated}]
    $ &
    $\displaystyle w_{\text{term}} = -10.0$ \\

    Joint acceleration penalty &
    $\displaystyle
    c_{\text{joint-acc}}=
    \|\ddot{\mathbf{q}}\|_2^2
    $ &
    $\displaystyle w_{\text{joint-acc}} = -3\times 10^{-7}$ \\

    Joint limit penalty &
    $\displaystyle
    c_{\text{joint-lim}}=
    \sum_{j=1}^{12}
    \Bigl[
    \max(\underline q^{\text{soft}}_j-q_j,0)
    +
    \max(q_j-\overline q^{\text{soft}}_j,0)
    \Bigr]
    $ &
    $\displaystyle w_{\text{joint-lim}} = -1.0$ \\

    Action rate penalty &
    $\displaystyle
    c_{\text{act-rate}}=
    \|\mathbf{a}_t-\mathbf{a}_{t-1}\|_2^2
    $ &
    $\displaystyle w_{\text{act-rate}} = -0.045$ \\

    Bad two-foot contact penalty &
    $\displaystyle
    c_{\text{bad-2}}=
    \chi\,
    \mathbbold{1}[|C|=2]\,
    \mathbbold{1}\!\left[
    C\notin
    \{\{\mathrm{FL},\mathrm{RR}\},\{\mathrm{FR},\mathrm{RL}\}\}
    \right]
    $ &
    $\displaystyle w_{\text{bad-2}} = -0.7$ \\

    Feet clearance penalty &
    $\displaystyle
    c_{\text{clr}}=
    \chi
    \sum_{i=1}^{4}
    |z_i-z_{\text{tar}}|\,\|\mathbf{v}_{i,xy}^{\text{foot}}\|_2,
    \qquad
    z_{\text{tar}}=0.07
    $ &
    $\displaystyle w_{\text{clr}} = -1.0$ \\

    Feet slip penalty &
    $\displaystyle
    c_{\text{slip}}=
    \chi
    \sum_{i=1}^{4}
    \|\mathbf{v}_{i,xy}^{\text{foot}}\|_2^2\,
    \mathbbold{1}[i\in\text{contact}]
    $ &
    $\displaystyle w_{\text{slip}} = -0.12$ \\

    Soft landing penalty &
    $\displaystyle
    c_{\text{land}}=
    \chi
    \sum_{i=1}^{4}
    f_i^{\text{contact}}\,
    \mathbbold{1}[i\in\text{first-contact}]
    $ &
    $\displaystyle w_{\text{land}} = -2\times 10^{-4}$ \\

    \bottomrule
    \end{tabularx}
    \vspace{2pt}
    \caption{Penalty terms for the fully reward-shaped quadruped velocity-tracking task. Here $C=\{i:I_i=1\}$ is the set of feet in contact.}
    \label{tab:full_reward_penalty}
\end{table}

\clearpage


\subsection{Converting Quadruped MPC Trajectories to RL Transitions}
\label{appendix:converting_quadruped_mpc_to_rl}

For the quadruped, the MPC outputs joint torques while the RL policy outputs joint position residuals, so we must convert each recorded MPC transition into an action-labeled RL transition. Let $k$ index control steps and $i \in \{0,\dots,D-1\}$ index the simulator substeps within a control step, with $D=4$ in our quadruped experiments. We denote by $\tau^{\mathrm{app}}_{k,i}$ the joint torque sent to the simulator at substep $(k,i)$, which the MPC-controlled system computes as
\begin{equation}
\tau^{\mathrm{app}}_{k,i} = \tau^{\mathrm{mpc}}_{k,i} + K_p^{\mathrm{mpc}}\!\left(q^{\mathrm{des}}_{k,i} - q_{k,i}\right) - K_d^{\mathrm{mpc}} \dot q_{k,i},
\end{equation}
where $q_{k,i}$ and $\dot q_{k,i}$ are the measured joint positions and velocities at substep $(k,i)$.

In the RL environment, the policy outputs a residual action $a_k$ that is converted into a joint target $q^{\mathrm{tgt}}_k = q^{\mathrm{nom}} + s\, a_k$, where $q^{\mathrm{nom}}$ is the nominal standing configuration and $s$ is the action scale. The target is held fixed across all $D$ substeps of the control interval and tracked by a PD controller:
\begin{equation}
\tau^{\mathrm{rl}}_{k,i} = K_p^{\mathrm{rl}}\!\left(q^{\mathrm{tgt}}_k - q_{k,i}\right) - K_d^{\mathrm{rl}} \dot q_{k,i}.
\end{equation}

To associate each MPC control interval with an RL action, we choose the target so that the RL torque matches the recorded applied torque at the first substep of the interval, $\tau^{\mathrm{rl}}_{k,0} = \tau^{\mathrm{app}}_{k,0}$. Solving for the joint target and the corresponding residual action gives
\begin{align}
q^{\mathrm{tgt}}_k &= q_{k,0} + \frac{\tau^{\mathrm{app}}_{k,0} + K_d^{\mathrm{rl}} \dot q_{k,0}}{K_p^{\mathrm{rl}}}, \label{eq:inverse_pd_target} \\
a_k &= \frac{q^{\mathrm{tgt}}_k - q^{\mathrm{nom}}}{s}, \label{eq:inverse_pd_action}
\end{align}
where all operations are applied elementwise across joints.

Because $q^{\mathrm{tgt}}_k$ is held fixed over the full control interval, the torque match is exact only at substep $i=0$. At later substeps, the state evolves to $(q_{k,i}, \dot q_{k,i})$, and the RL controller would produce
\begin{equation}
\tau^{\mathrm{rl}}_{k,i} = \tau^{\mathrm{app}}_{k,0} + K_p^{\mathrm{rl}}\!\left(q_{k,0} - q_{k,i}\right) + K_d^{\mathrm{rl}}\!\left(\dot q_{k,0} - \dot q_{k,i}\right),
\end{equation}
giving a substep-wise torque mismatch under standard action replay of
\begin{equation}
\Delta \tau_{k,i} = \tau^{\mathrm{rl}}_{k,i} - \tau^{\mathrm{app}}_{k,i} = \tau^{\mathrm{app}}_{k,0} - \tau^{\mathrm{app}}_{k,i} + K_p^{\mathrm{rl}}\!\left(q_{k,0} - q_{k,i}\right) + K_d^{\mathrm{rl}}\!\left(\dot q_{k,0} - \dot q_{k,i}\right).
\label{eq:substep_mismatch}
\end{equation}

If the target could be recomputed at every simulator substep, the exact target would be
\begin{equation}
q^{\mathrm{tgt},*}_{k,i} = q_{k,i} + \frac{\tau^{\mathrm{app}}_{k,i} + K_d^{\mathrm{rl}} \dot q_{k,i}}{K_p^{\mathrm{rl}}},
\label{eq:per_substep_target}
\end{equation}
which would satisfy $\tau^{\mathrm{rl}}_{k,i} = \tau^{\mathrm{app}}_{k,i}$ for every substep $i$.

Because this conversion is exact only at the first simulator substep of each control interval, the injected quadruped transitions are approximate off-policy samples rather than exact ones; we retain the standard injection pipeline and treat the residual action mismatch as a modeling approximation. Across the collected data, action clipping occurred in fewer than 1\% of environment steps.


\section{Why Replay-Distribution Bias Selects Behaviors}
\label{appendix:why_mpc_rl_works}

We provide an intuition-level account of how MPC-Injection biases behavior selection in replay-buffer-based actor-critic algorithms. We do not claim convergence guarantees under function approximation. The purpose of the analysis is to explain why MPC-Injection consistently drives the policy into the controller's behavior basin under simple task rewards where vanilla RL converges elsewhere.

\textbf{Distribution mismatch and the critic.} A natural concern is whether injecting MPC transitions breaks the off-policy learning machinery. Consider the actor gradient for TD3,
\begin{equation} \label{eq:td3_policy_grad}
    \nabla_\theta \frac{1}{|B|} \sum_{s \in B} Q_{\phi_1}(s,\pi_\theta(s)),
\end{equation}
where $B$ is a batch of transitions from the replay buffer $\mathcal{D}$ and $Q_{\phi_i}$ are the two learned Q-functions with parameters $\phi_i$ for $i=1,2$. The analogous actor gradient for SAC is
\begin{equation} \label{eq:sac_policy_grad}
    \nabla_\theta \frac{1}{|B|} \sum_{s \in B} \left( \min_{i=1,2} Q_{\phi_i}(s,\tilde{a}_\theta(s)) - \alpha \log \pi_\theta(\tilde{a}_\theta(s) | s) \right),
\end{equation}
where $\tilde{a}_\theta(s)$ is a reparameterized sample from $\pi_\theta(\cdot|s)$. The corresponding critic update is
\begin{equation} \label{eq:critic_update}
    \nabla_{\phi_i} \frac{1}{|B|} \sum_{(s,a,r,s') \in B} (Q_{\phi_i}(s,a) - y(r,s'))^2, \quad i=1,2.
\end{equation}

The actor and critic use the same replay minibatch differently. Injecting MPC transitions changes the states $s$ at which the actor evaluates current-policy actions, while the recorded $(s,a,r,s')$ tuples directly affect critic training. Because the rollouts are executed in the RL simulator, this is compatible with off-policy learning from a mixture of behavior policies. For the quadruped, however, the inverse-PD action conversion is exact only at the first simulator substep, so those injected tuples retain the approximation described in Appendix~\ref{appendix:converting_quadruped_mpc_to_rl}.

\textbf{Where behavior selection comes from.} The critic remaining valid does not by itself explain why $\pi_\theta$ converges to the controller's behavior basin rather than simply learning a more accurate $Q$ over a wider state distribution. The behavior-selection effect arises from the actor objective. Both Eqs.~\ref{eq:td3_policy_grad} and \ref{eq:sac_policy_grad} evaluate $Q$ at states sampled from $\mathcal{D}$ and at actions produced by the \emph{current policy}. Without MPC-Injection, $\mathcal{D}$ is filled with states the current policy reaches during its own rollouts, so the actor is optimized only over the basin the policy currently occupies. If early training drives the policy into an undesirable basin, such as the walker's scooting behavior, the actor is updated exclusively on scooting states, the policy stays there, and the basin reinforces itself.

\begin{figure}[t]
  \centering
  \includegraphics[width=0.5\linewidth]{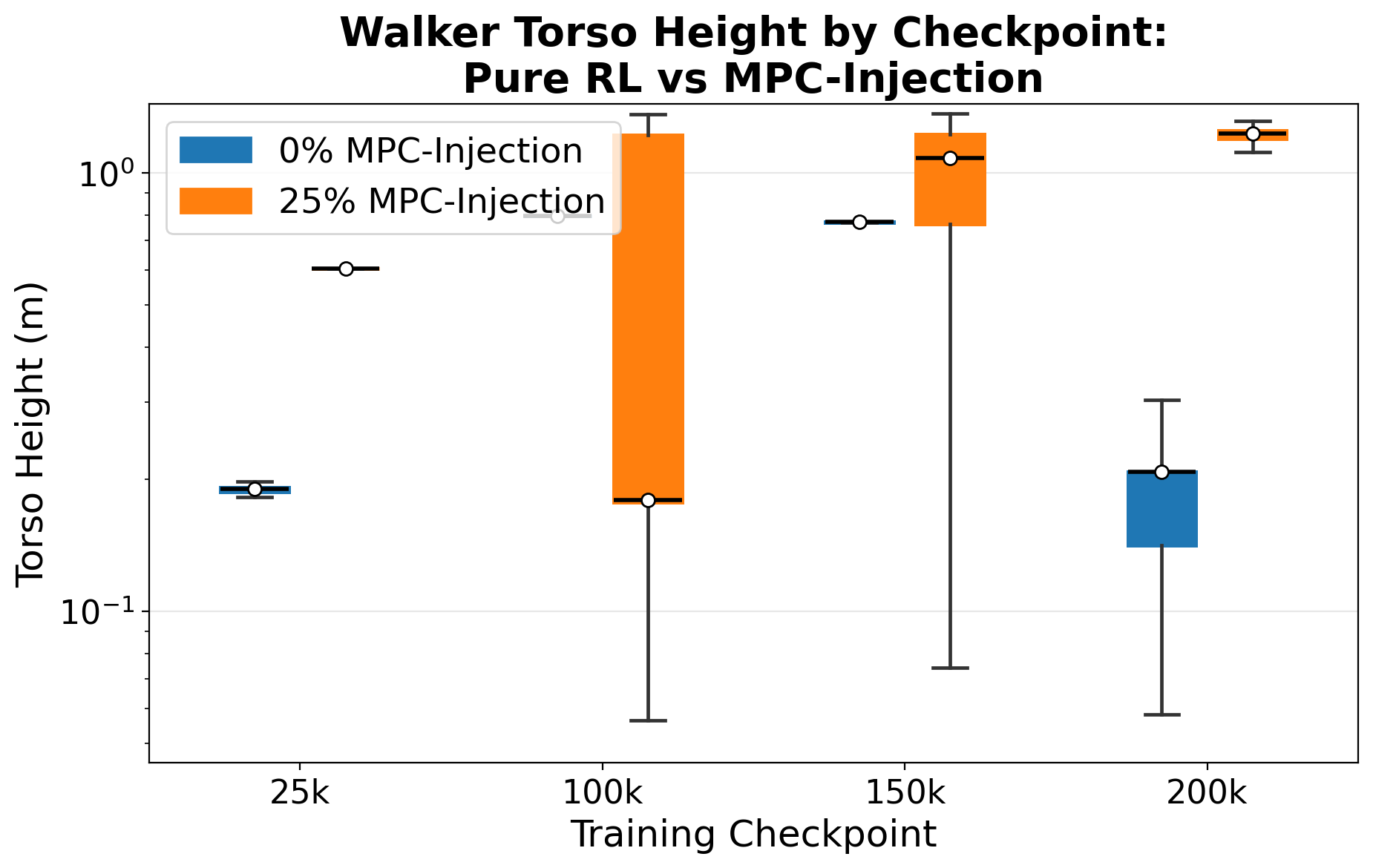}
  \caption{Log-scale box plot showing the evolution of torso-height distributions at different training checkpoints for 0\% and 25\% MPC-Injection, where MPC-Injection biases learning toward the gait demonstrated by MPC transitions and leads to earlier emergence of upright locomotion under the same reward.}
  \label{fig:torso_traj_over_training}
\end{figure}

\begin{table}[t]
  \centering
  \caption{Torso-height behavior of walker policies during training. Values summarize the qualitative torso-height distributions shown in Fig.~\ref{fig:torso_traj_over_training}. Pure RL eventually collapses to a low-torso strategy, while 25\% MPC-Injection converges to an upright locomotion behavior.}
  \vspace{2pt}
  \label{tab:walker_torso_height}
  \small
  \renewcommand{\arraystretch}{1.15}
  \setlength{\tabcolsep}{4pt}
  \begin{tabularx}{\linewidth}{
    >{\centering\arraybackslash}p{0.14\linewidth}
    >{\centering\arraybackslash}p{0.22\linewidth}
    >{\centering\arraybackslash}p{0.27\linewidth}
    >{\raggedright\arraybackslash}X
  }
    \toprule
    \textbf{Checkpoint} &
    \textbf{Pure RL torso height} &
    \textbf{25\% MPC-Injection torso height} &
    \textbf{Main takeaway} \\
    \midrule

    25k &
    Low, around 0.19 m &
    Around 0.60 m &
    Pure RL begins in a low-torso height regime. \\

    100k &
    Around 0.80 m &
    Highly variable, 0.06--1.36 m &
    MPC-Injection begins exploring upright postures but has not stabilized. \\

    150k &
    Around 0.77 m &
    Median around 1.08 m &
    MPC-Injection shifts the learned behavior toward upright walking. \\

    200k &
    Low, around 0.20 m &
    Stable upright, around 1.23 m &
    MPC-Injection converges to upright locomotion while pure RL collapses to a low-torso strategy. \\

    \bottomrule
  \end{tabularx}
\end{table}

Figure~\ref{fig:torso_traj_over_training} and Table~\ref{tab:walker_torso_height} show this self-confining dynamic empirically. With $p=0$ (vanilla RL), the walker's torso height collapses to ground-scooting values. With $p=25\%$ MPC-Injection, the torso height stays in the upright range early on in training. The basin separation appears relatively early and persists, consistent with the actor being optimized over different state distributions in the two cases. We note that the pure RL torso height distribution in early training is due to both random exploration and random initialization. After a few timesteps in an episode, the pure RL policy falls, leading to the learned scooting behavior. By contrast, MPC-Injection learns a walking gait early in training.

\textbf{The injection mechanism.} MPC-Injection breaks the self-confining loop by placing MPC-visited states in $\mathcal{D}$. The actor is then updated to take high-$Q$ actions at states the current policy would otherwise never see, such as the upright periodic configurations along the MPC trajectory. Crucially, the actor's update at these states uses its own action $\pi_\theta(s)$, not the recorded MPC action. This distinguishes MPC-Injection from behavior cloning: the policy is not asked to match the MPC's actions at MPC-visited states, but to take whatever action maximizes the learned $Q$ at those states. The critic, trained on rewards and dynamics observed along both MPC and on-policy trajectories, supplies the $Q$-values that make this evaluation meaningful. Over training, the combined update drives $\pi_\theta$ toward actions whose trajectories overlap with the controller's basin while still allowing on-policy refinement within that basin.

\textbf{Role of the injection ratio.} This view clarifies why the injection ratio $p$ matters. If $p$ is too small, MPC-visited states are rarely sampled in a batch and the actor gradient is dominated by states from the policy's current basin, recovering vanilla RL. If $p$ is too large, the actor gradient is dominated by states drawn from a distribution the policy cannot consistently reach via its own rollouts, and the policy never closes the loop between the basin pull and its own visitation distribution. The analysis therefore predicts an intermediate regime in which the actor is updated frequently enough on MPC-visited states to be pulled toward the controller's basin and frequently enough on its own rollouts to consolidate that basin as a region it can reach and stabilize from. Section~\ref{sec:behavior_biasing_under_and_identical_reward} finds that $p=25\%$ is such a regime, and Figure~\ref{fig:walkers_diff_pct} shows that this is the ratio at which the walker's gait is most structured, while higher ratios produce less coherent gaits.

\textbf{A predicted failure mode.} Finally, this view predicts a specific failure mode under extended training. The basin pull from MPC-Injection is implicit and acts only through the actor's evaluation of $Q$ at MPC-visited states. Once the policy operates inside the controller's basin, the on-policy rollouts entering $\mathcal{D}$ overlap heavily with the injected MPC states, and the additional pull contributed by the MPC transitions diminishes. The actor objective is then approximately what it would be under vanilla RL restricted to the controller's basin, with no explicit term anchoring it there. Behaviors that improve local return without preserving the basin, such as lower torso height, would therefore not be penalized, and the policy may drift over extended training. This is in contrast to reward shaping or AMP, in which the corresponding term remains in the reward or critic signal throughout training and continues to shape $Q$ and $\pi_\theta$ toward the desired basin.

\section{Ablation on Increased MPC-Injection Percentages} \label{appendix:mpc-injection_pct_ablation}

\begin{figure*}[!t]
  \centering
  \setlength{\tabcolsep}{4pt}
  \renewcommand{\arraystretch}{0}
  \newcommand{\colw}{0.45\textwidth}
  \newcommand{\V}[1]{%
    \includegraphics[width=\colw, valign=t]{figures/walker/pct_behaviors/viser_sac_mpc_#1pct_labeled.png}%
  }
  \newcommand{\R}[1]{%
    \includegraphics[
      width=\colw,
      valign=t,
      trim=0bp 316.5bp 0bp 0bp,
      clip
    ]{figures/walker/raster_footstep_plots/raster_ftstp_plots_sac_mpc_#1pct_500000.png}%
  }

  \begin{tabular}{@{}cc@{}}
    \V{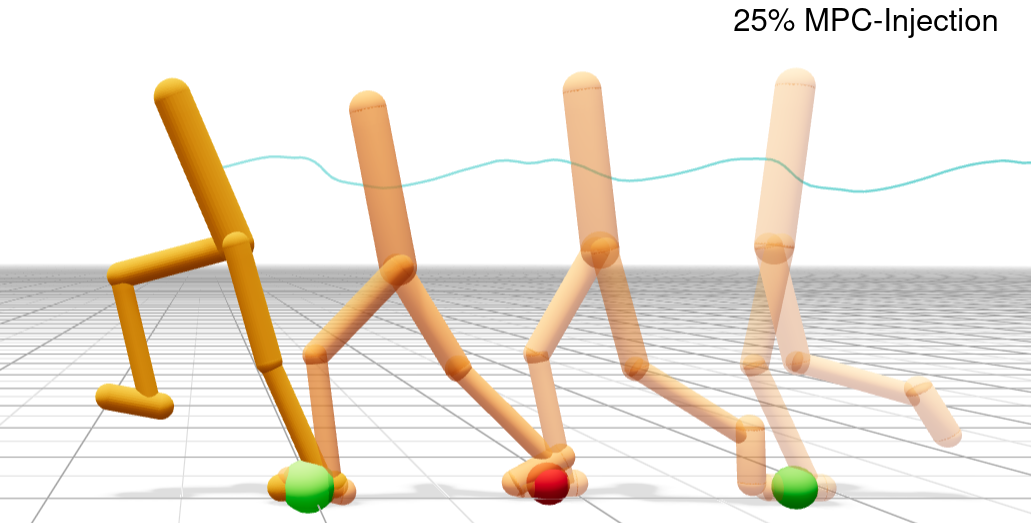} & \V{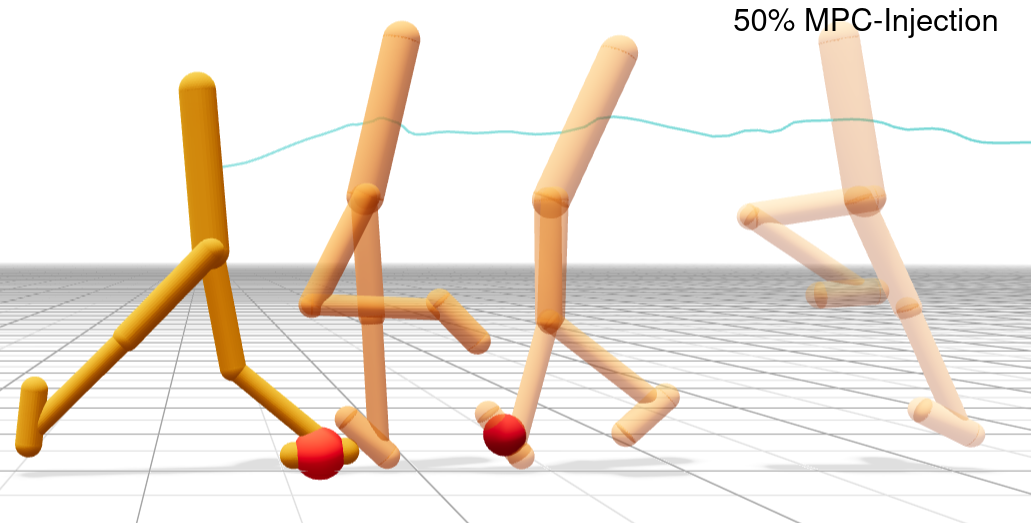} \\[-2pt]
    \R{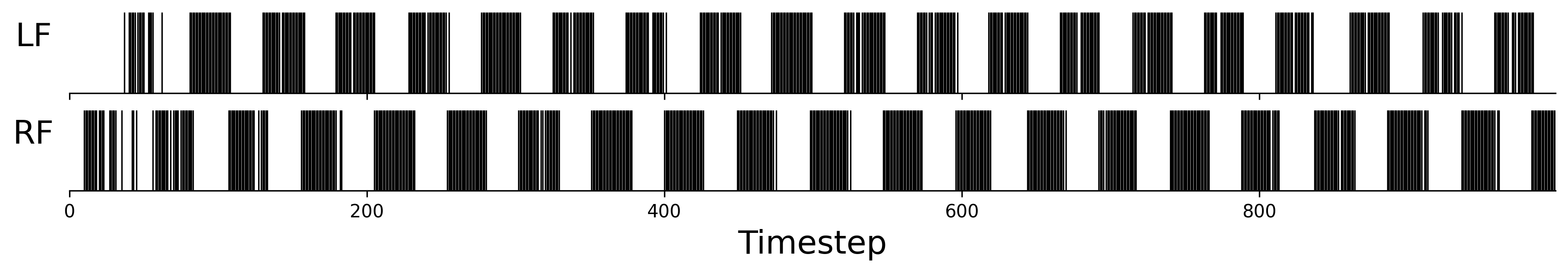} & \R{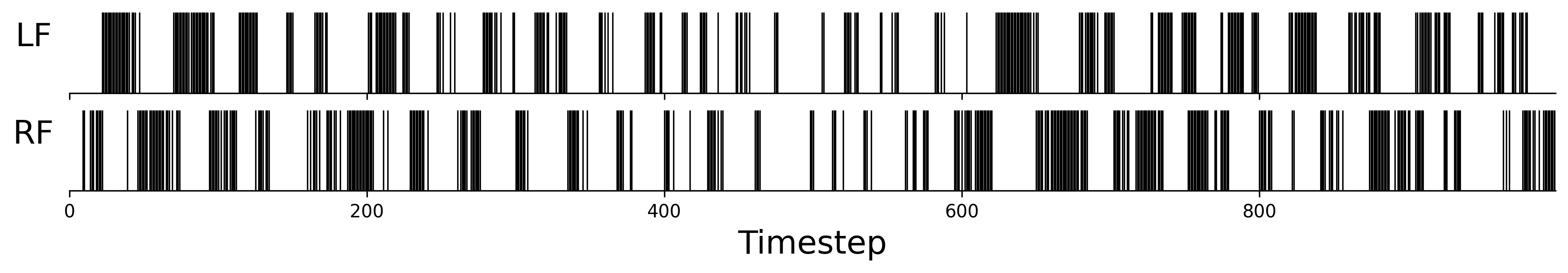} \\[4pt]
    \V{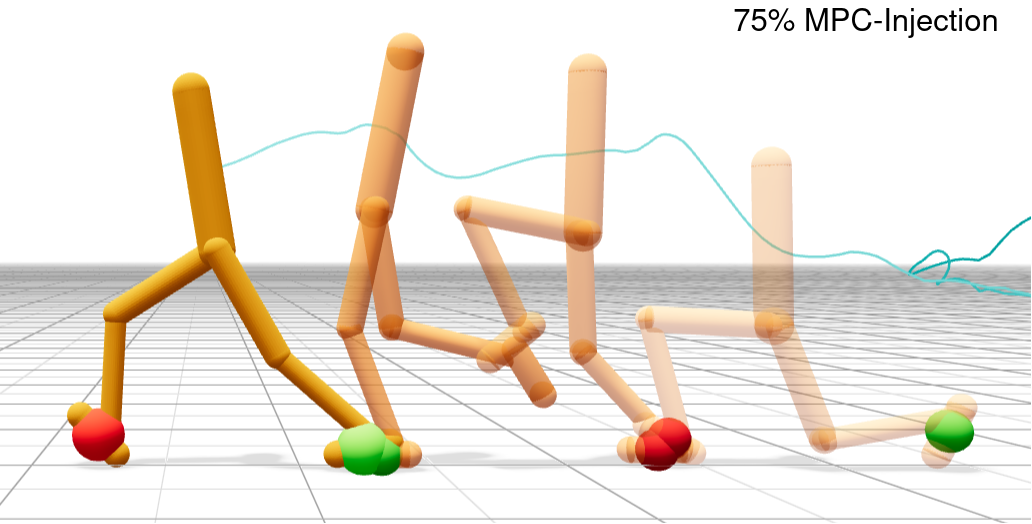} & \V{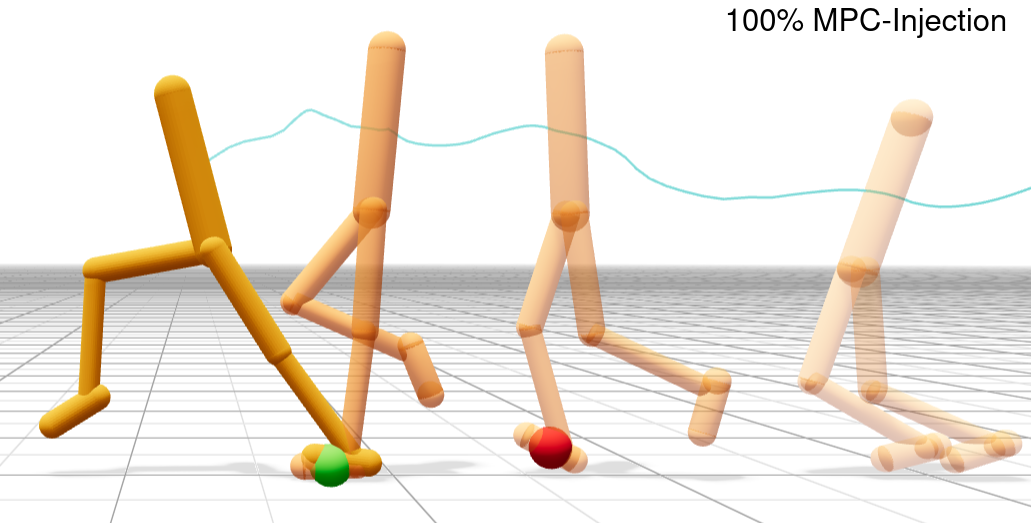} \\[-2pt]
    \R{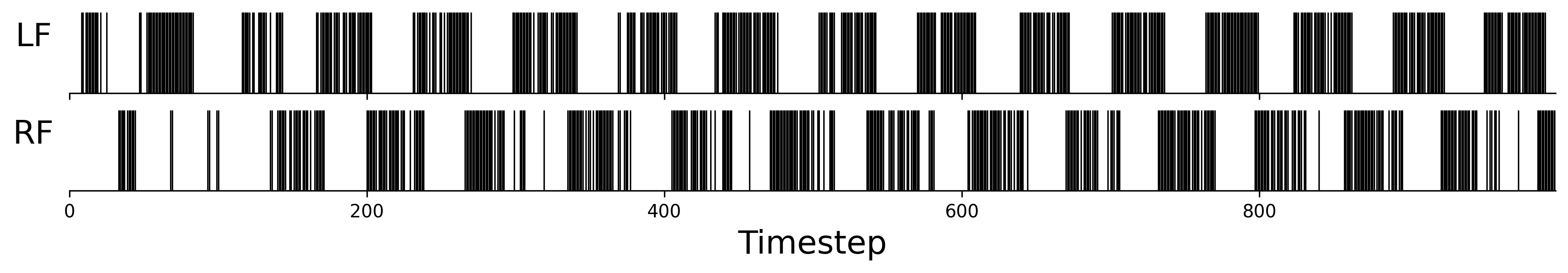} & \R{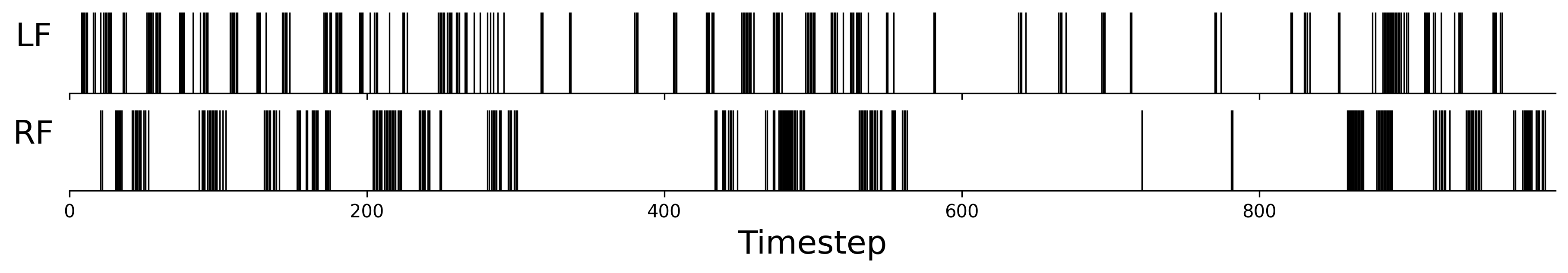}
  \end{tabular}
  \caption{Walker behavior under varying MPC-Injection ratios (25\% top-left, 50\% top-right, 75\% bottom-left, 100\% bottom-right). Each panel shows a walking-behavior snapshot. As the injection ratio increases above 25\%, the behavior becomes more erratic and diverges from the MPC behavior.}
  \label{fig:walkers_diff_pct}
\end{figure*}

\textbf{2D Walker.} Figure~\ref{fig:walkers_diff_pct} shows the walker's behavior as the injection ratio increases above 25\%. At 50\% and 75\%, the policy continues to achieve high episodic return (Figure~\ref{fig:learning_rate_sac_td3_mpc}) but produces erratic footstep patterns, including jumping and falling. At 100\%, the policy fails to consistently produce a stable gait and episodic return drops as well. These results are consistent with the analysis in Appendix~\ref{appendix:why_mpc_rl_works}. As the injection ratio increases, the actor gradient becomes dominated by states the policy cannot consistently reach via its own rollouts, preventing it from reaching any single basin.

\begin{figure}[!t]
    \centering
    \includegraphics[width=0.55\linewidth]{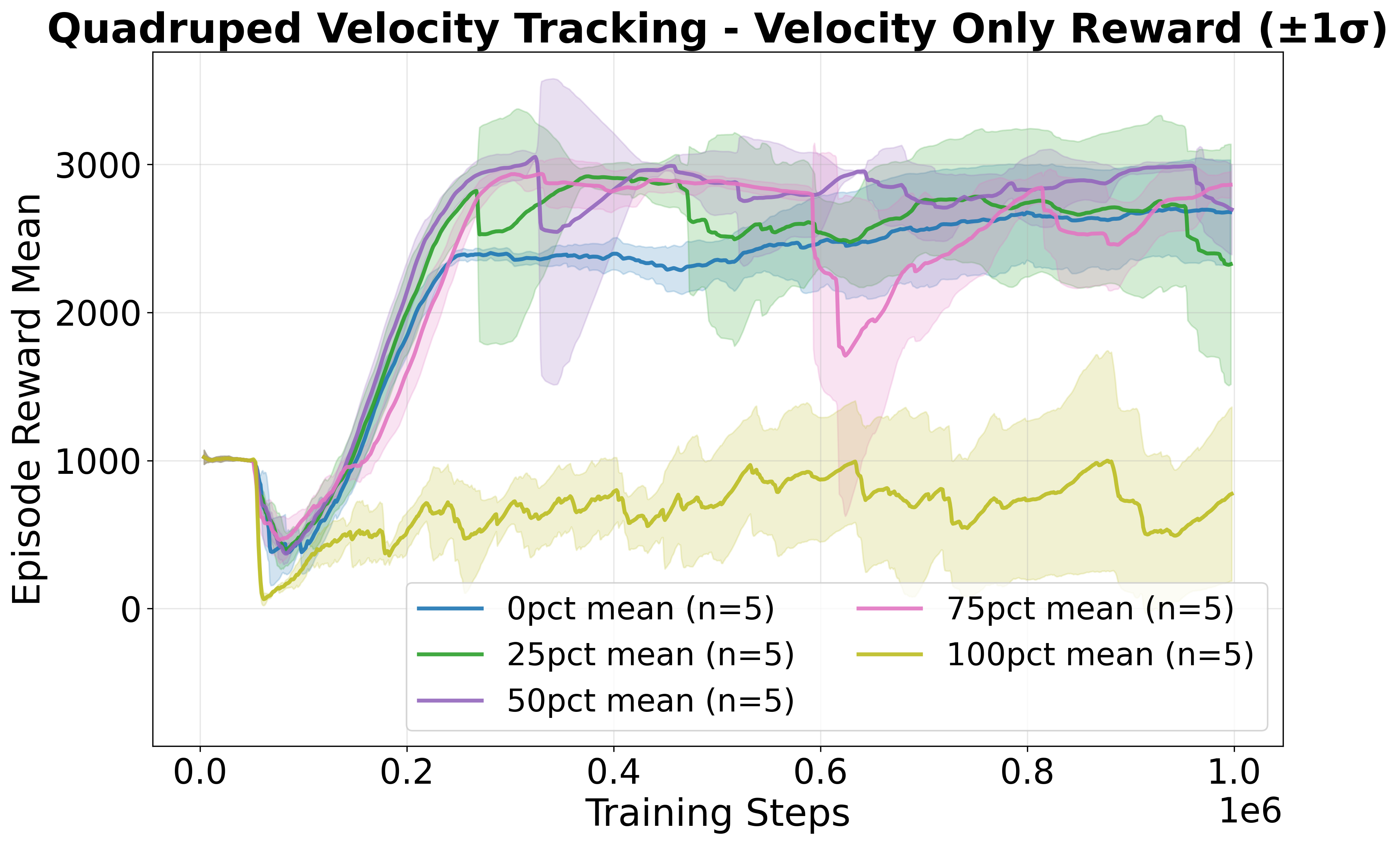}
    \caption{Quadruped training curves across 0\%, 25\%, 50\%, 75\%, and 100\% MPC-Injection for SAC. Behavior convergence emerges around 400k timesteps; on a subset of seeds, drift was observed after extended training even though reward remains high (see Appendix~\ref{appendix:why_mpc_rl_works} for the predicted mechanism).}
    \label{fig:quad_training_curves}
\end{figure}

\textbf{Quadruped.} The quadruped showed similar results to the walker, with injection percentages higher than 25\% leading to degraded behavior. Figure~\ref{fig:quad_training_curves} presents the corresponding training curves, where behavior convergence to the MPC basin emerges early at around 400k timesteps. On a subset of seeds we observed behavior drift after extended training, consistent with the prediction in Appendix~\ref{appendix:why_mpc_rl_works}, though further experiments are needed to confirm this is a general phenomenon rather than simply a seed-specific effect.

\section{Hardware and Sim-to-Real Transfer} \label{appendix:sim2real}

\textbf{Robot system.} The quadruped is a 12-DOF Unitree Go2. The RL policy runs on a tethered desktop computer (Intel Core Ultra 9 CPU, 64 GB RAM, NVIDIA GeForce RTX 5090) that communicates with the Go2 over Ethernet.

\textbf{Simulation-side preparation.} We use the domain-randomization parameters in Table~\ref{tab:domain-randomization} for MPC-Injection, reward shaping, and the evaluated AMP-MPC baseline. We additionally calibrate per-joint damping, armature, and friction-loss using MuJoCo's system identification utilities to bring the simulation closer to the physical Go2.

\textbf{Deployment stack.} A low-pass filter sits between the policy output and the low-level PD controller, with a cutoff of 5.0~Hz. The PD controller tracks joint position targets with gains $K_p^{\mathrm{rl}}=20$ and $K_d^{\mathrm{rl}}=1$ for the hip and thigh joints, while the calf joints have gains $K_p^{\mathrm{rl}}=40$ and $K_d^{\mathrm{rl}}=2$. The evaluated reward-shaped, MPC-Injection, and AMP-MPC policies share the same low-pass filter and PD gains. The reported hardware results are qualitative and do not isolate policy effects from all other deployment choices.

\begin{table}[h]
    \centering
    \small
    \setlength{\tabcolsep}{4pt}
    \caption{Domain-randomization parameters shared by MPC-Injection, reward shaping, and the evaluated AMP-MPC baseline.
    }
    \label{tab:domain-randomization}
    \begin{tabular}{@{}lll@{}}
    \toprule
    Category & Parameter & Range / Value \\
    \midrule
    Contact & Tangential friction & $[0.3,\,1.6]$ \\
    Base dynamics & Added base mass & $[-1.0,\,3.0]$ kg \\
    Base dynamics & CoM displacement, each axis & $[-0.05,\,0.05]$ m \\
    Sensing & Joint encoder bias & $[-0.015,\,0.015]$ rad \\
    Control & $K_p$ scale & $[0.9,\,1.1]$ \\
    Control & $K_d$ scale & $[0.9,\,1.1]$ \\
    Joint dynamics & Damping scale & $[0.8,\,1.2]$ \\
    Joint dynamics & Armature scale & $[0.8,\,1.2]$ \\
    Joint dynamics & Friction-loss scale & $[0.8,\,1.2]$ \\
    Actuation & Motor strength scale & $[0.9,\,1.1]$ \\
    Observation noise & Joint position & $\pm 0.01$ rad \\
    Observation noise & Joint velocity & $\pm 1.5$ rad/s \\
    Observation noise & Angular velocity & $\pm 0.2$ rad/s \\
    Observation noise & Gravity estimate & $\pm 0.05$ \\
    Pushes & Interval & $[5.0,\,10.0]$ s \\
    Pushes & Linear velocity kick $(x,y,z)$ & $[-0.5,0.5]$, $[-0.5,0.5]$, $[-0.25,0.25]$ m/s \\
    Pushes & Angular velocity kick $(r,p,y)$ & $[-0.52,0.52]$, $[-0.52,0.52]$, $[-0.78,0.78]$ rad/s \\
    \bottomrule
    \end{tabular}
\end{table}

\end{document}